\documentclass[lettersize,journal]{IEEEtran}
\usepackage{amsmath,amsfonts}
\usepackage{algorithmic}
\usepackage{algorithm}
\usepackage{array}
\usepackage[caption=false,font=footnotesize,labelfont=rm,textfont=rm]{subfig}
\usepackage{textcomp}
\usepackage{stfloats}
\usepackage{url}
\usepackage{verbatim}
\usepackage{graphicx}
\usepackage{cite}
\usepackage{amsmath}
\usepackage{booktabs}
\usepackage{comment}
\usepackage{multicol,multirow}
\usepackage{color}
\usepackage[dvipsnames]{xcolor,colortbl}
\usepackage{makecell}
\usepackage[accsupp]{axessibility}
\usepackage{threeparttable}
\usepackage{marvosym}
\usepackage{tabularx}

\usepackage[colorlinks,
            breaklinks,
            pagebackref,
            linkcolor=blue,
            anchorcolor=blue,
            citecolor=blue
            ]{hyperref}

\usepackage{xcolor}

\newcommand{\ie}{\textit{i}.\textit{e}.}

\newcommand{\etal}{\textit{et}~\textit{al}.}

\begin{document}

\title{SkeletonX: Data-Efficient Skeleton-based Action Recognition via Cross-sample Feature Aggregation}

\author{Zongye~Zhang,
            Wenrui~Cai,
            Qingjie~Liu,~\IEEEmembership{Member,~IEEE,}
            Yunhong~Wang,~\IEEEmembership{Fellow,~IEEE}
\thanks{Corresponding author: Qingjie~Liu.}
\thanks{Zongye Zhang, Wenrui Cai, Qingjie Liu, and Yunhong Wang are with the State Key Laboratory of Virtual Reality Technology and Systems, Beihang University, Beijing, China.	E-mail: \{zhangzongye, wenrui\_cai, qingjie.liu, yhwang\}@buaa.edu.cn}
\thanks{This research was supported by the “Pioneer” and “Leading Goose” R\&D Program of Zhejiang (No. 2024C01020), and the National Natural Science Foundation of China under grants 62176017.}
\thanks{© 2025 IEEE.  Personal use of this material is permitted.  Permission from IEEE must be obtained for all other uses, in any current or future media, including reprinting/republishing this material for advertising or promotional purposes, creating new collective works, for resale or redistribution to servers or lists, or reuse of any copyrighted component of this work in other works.}
}

\maketitle
\begin{abstract}
While current skeleton action recognition models demonstrate impressive performance on large-scale datasets, their adaptation to new application scenarios remains challenging. These challenges are particularly pronounced when facing new action categories, diverse performers, and varied skeleton layouts, leading to significant performance degeneration. Additionally, the high cost and difficulty of collecting skeleton data make large-scale data collection impractical. This paper studies one-shot and limited-scale learning settings to enable efficient adaptation with minimal data. Existing approaches often overlook the rich mutual information between labeled samples, resulting in sub-optimal performance in low-data scenarios. To boost the utility of labeled data, we identify the variability among performers and the commonality within each action as two key attributes. We present SkeletonX, a lightweight training pipeline that integrates seamlessly with existing GCN-based skeleton action recognizers, promoting effective training under limited labeled data. First, we propose a tailored sample pair construction strategy on two key attributes to form and aggregate sample pairs. Next, we develop a concise and effective feature aggregation module to process these pairs. Extensive experiments are conducted on NTU RGB+D, NTU RGB+D 120, and PKU-MMD with various GCN backbones, demonstrating that the pipeline effectively improves performance when trained from scratch with limited data. Moreover, it surpasses previous state-of-the-art methods in the one-shot setting, with only 1/10 of the parameters and much fewer FLOPs. The code and data are available at: https://github.com/zzysteve/SkeletonX
\end{abstract}


\IEEEpeerreviewmaketitle

\section{Introduction}
\label{sec:intro}

\IEEEPARstart{H}{uman} action recognition has been a fast-developing field in recent years and can be applied to human-machine interaction, virtual reality, video understanding, \textit{etc}.
Different from RGB videos, skeleton sequences exclusively capture the coordinates of key joints, which are robust against variations in camera viewpoints, background, and appearance. With the development of depth sensor and human pose estimation methods\cite{pavlakosCoarseToFineVolumetricPrediction2017, caiExploitingSpatialTemporalRelationships2019,zhaoSemanticGraphConvolutional2019, zhao2024crosscamera}, several large-scale skeleton-based datasets~\cite{shahroudyNTURGBLarge2016, liuNTURGB1202020,liuBenchmarkDatasetComparison2020} are proposed. Graph Convolutional Networks (GCNs) have gained significant attention in skeleton-based action recognition, as human skeletons are naturally graph-structured data. Numerous methods~\cite{yanSpatialTemporalGraph2018,shiTwoStreamAdaptiveGraph2019,liuDisentanglingUnifyingGraph2020,chengDecouplingGCNDropGraph2020,korbanDDGCNDynamicDirected2020,chenChannelwiseTopologyRefinement2021,wangSkeletonbasedActionRecognition2022,leeHierarchicallyDecomposedGraph2023, zhouBlockGCNRedefineTopology2024, zhangCrossScaleSpatiotemporalRefinement2024} have been proposed for skeleton-based action recognition, achieving impressive performance.

\begin{figure*}[t]
  \centering
  \includegraphics[width=\linewidth]{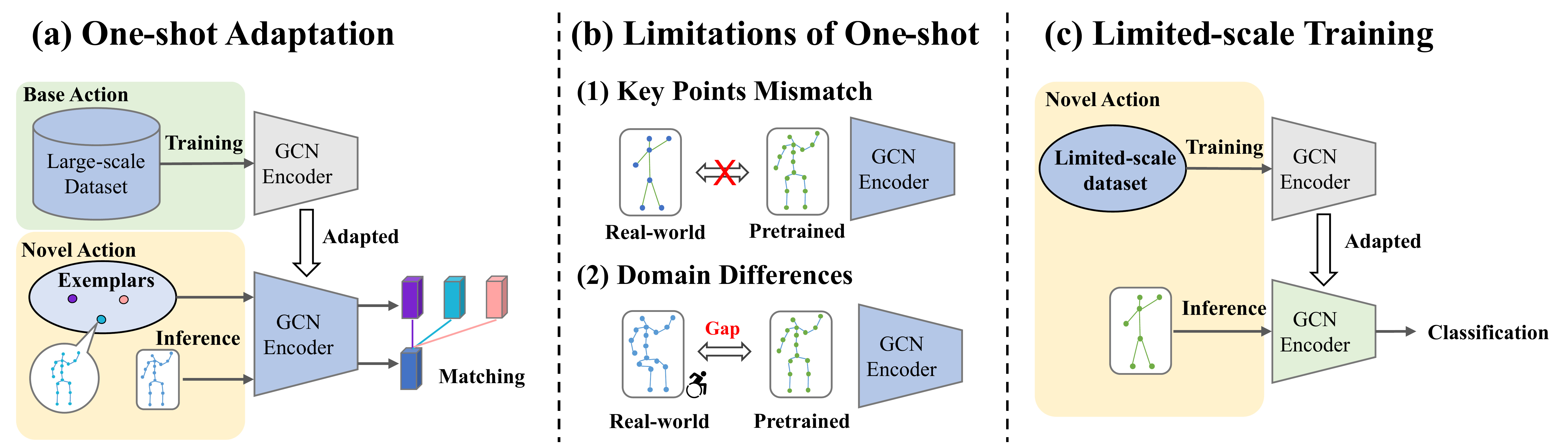}
  \caption{Illustration of two adaptation paradigms. (a) In the one-shot setting, feature encoders are first pretrained on large-scale datasets. When adapting to the inference phase, novel samples are recognized by matching them with features extracted by the pre-trained encoder. (b) However, this pretrain-and-matching paradigm fails when the number of key points differs from the pretraining phase. Even if the joint layouts are the same, domain gaps can occur, which degrade the model's performance. (c) To address this, we propose a limited-scale setting where the model is trained effectively from scratch with minimal newly collected data. This approach adapts the model directly to the inference phase without matching.}
\label{fig:two_settings}
\end{figure*}

Despite these advancements, most existing approaches rely heavily on large-scale, high-quality labeled datasets, which hinders adaptability in novel application scenarios. Creating such datasets requires advanced depth sensors and extensive human effort~\cite{liuNTURGB1202020, wang2024survey}, making the process costly and impractical. To mitigate these challenges, recent studies have shifted toward one-shot skeleton-based action recognition~\cite{memmesheimerSkeletonDMLDeepMetric2022, wangTemporalViewpointTransportationPlan2022,yangOneShotActionRecognition2024}, where the novel actions are classified with only one reference per class. This paradigm follows a pretrain-and-match protocol: models are first pretrained on base action categories with abundant samples and then classify novel categories by matching features against the reference samples, as illustrated in Fig.~\ref{fig:two_settings}(a). One-shot recognition significantly reduces data requirements during inference, making it particularly suitable for real-world applications with limited annotated data.

However, deploying these one-shot models in real-world scenarios presents additional challenges, as illustrated in Fig.~\ref{fig:two_settings}. Human actions, represented in skeleton modality, are closely tied to skeleton modeling methods, including the number of joints and the skeletal layout. Different motion capture systems and human pose estimation methods~\cite{shahroudyNTURGBLarge2016, caoRealtimeMultiPerson2D2017} often generate skeletons with varying layouts and numbers of joints~\cite{kayKineticsHumanAction2017,liuNTURGB1202020}, leading to mismatches between the base actions in the pretraining phase and the novel actions in the inference phase. This discrepancy presents a significant limitation for GCN models, whose structures are inherently tied to the predefined skeleton topology. Consequently, models pretrained on large-scale datasets cannot be directly applied to new skeleton layouts. Moreover, performance may further degrade in domains with substantial distributional shifts, such as medical applications~\cite{sabaterOneShotActionRecognition2021}.

\begin{figure}[t]
  \centering
  \includegraphics[width=\columnwidth]{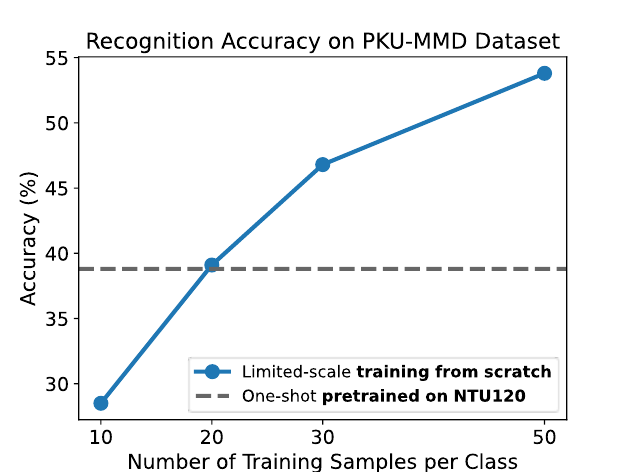}
  \caption{Illustration of the performance degeneration when transferring one-shot pretrained models on NTU RGB+D 120 dataset to PKU-MMD datasets. The gray dashed line indicates the performance of the one-shot setting. The blue line demonstrates the limited-scale setting, which requires collecting only a handful of samples and training the model \textbf{from scratch}. By collecting 20 samples per class for novel actions, the model achieves better performance than the one-shot pretrained one, and still benefits from continuous data accumulation.}
\label{fig:domain_gap}
\end{figure}

Therefore, we explore a novel and challenging limited-scale setting that addresses aspects that one-shot learning cannot fully resolve. Instead of adapting models through matching, the limited-scale paradigm aims to effectively train models \textbf{from scratch} using limited training samples, as illustrated in Fig.~\ref{fig:two_settings}(c). In this paradigm, the model is trained on a few samples collected from minimum subjects and camera setups collected specific to the target scenarios. Limited-scale training harnesses the performance benefits of continuous data collection while avoiding extra computational costs during inference, which are often associated with matching-based few-shot learning methods. As illustrated in Fig.~\ref{fig:domain_gap}, our method can achieve better performance by collecting 20 samples per class for novel actions and continues to improve as more data are collected. This setting also bridges the gap between one-shot/few-shot learning and large-scale training, where the former typically relies on no more than five samples per category, while the latter requires hundreds of annotated examples.

A key challenge in limited-scale settings arises from the inherent variability in actions performed by different individuals. Variations in body shapes, motion styles, and personal habits introduce significant diversity to action sequences, complicating recognition. As samples of novel action categories are scarce, the network struggles to learn diverse action representations from limited subjects, leading to degenerated feature extraction capabilities. Existing methods are designed for large-scale datasets, enhancing sample diversity through random rotation~\cite{memmesheimerSkeletonDMLDeepMetric2022, deyzelOneshotSkeletonbasedAction2023} and mixup~\cite{zhangMixupEmpiricalRisk2018, mengSampleFusionNetwork2019, zhang2025skeletonmix}, while the rich mutual information among different action sequences is overlooked.

To address these challenges, we propose a novel cross-sample aggregation framework, \textbf{SkeletonX}. Our approach targets two key attributes in skeleton-based action recognition: the variability among \textit{performers} and commonality within each \textit{action}. To leverage these attributes, we introduce a tailored sample-pair construction strategy that forms pair batches for the original input batch. Each sample is first encoded and processed through a disentanglement module. The disentangled features are aggregated by our proposed cross-sample feature aggregation module, effectively introducing diversity into the training process despite the limited availability of labeled samples. Additionally, we propose an action-aware aggregation loss to further enhance the learning process, which encourages the model to learn discriminative feature representation, while ensuring end-to-end training for the pipeline. SkeletonX is designed to be practical and flexible, allowing seamless integration into various GCN architectures. We also provide an analysis based on Information Bottleneck Theory~\cite{tishby2000information, alemi2017deep} to better understand the mechanism behind performance gain through cross-sample feature aggregation. Extensive experiments conducted on limited-scale and one-shot settings with three datasets and multiple backbones demonstrate the effectiveness and generalizability of our method.

Our contributions can be summarized as follows.

\begin{itemize}
    \item We introduce a sample pairs selection strategy that guides the model to capture distinctions and correlations between paired samples during training, focusing on performer and action attributes.
    \item We propose SkeletonX, a cross-sample aggregation pipeline with a lightweight plug-and-play module compatible with most GCN-based models, enhancing computational efficiency for practical inference.
    \item We explore two data-efficient scenarios, demonstrating that our method achieves state-of-the-art performance in one-shot settings and delivers improved results across multiple backbones on NTU RGB+D, NTU RGB+D 120, and PKU-MMD datasets.
    \item We provide an analysis for our proposed method using Information Bottleneck Theory, offering a theoretical foundation for extending the scope and effectiveness of our method in future applications.
\end{itemize}

\begin{figure*}[t]
  \centering
  \includegraphics[width=\textwidth]{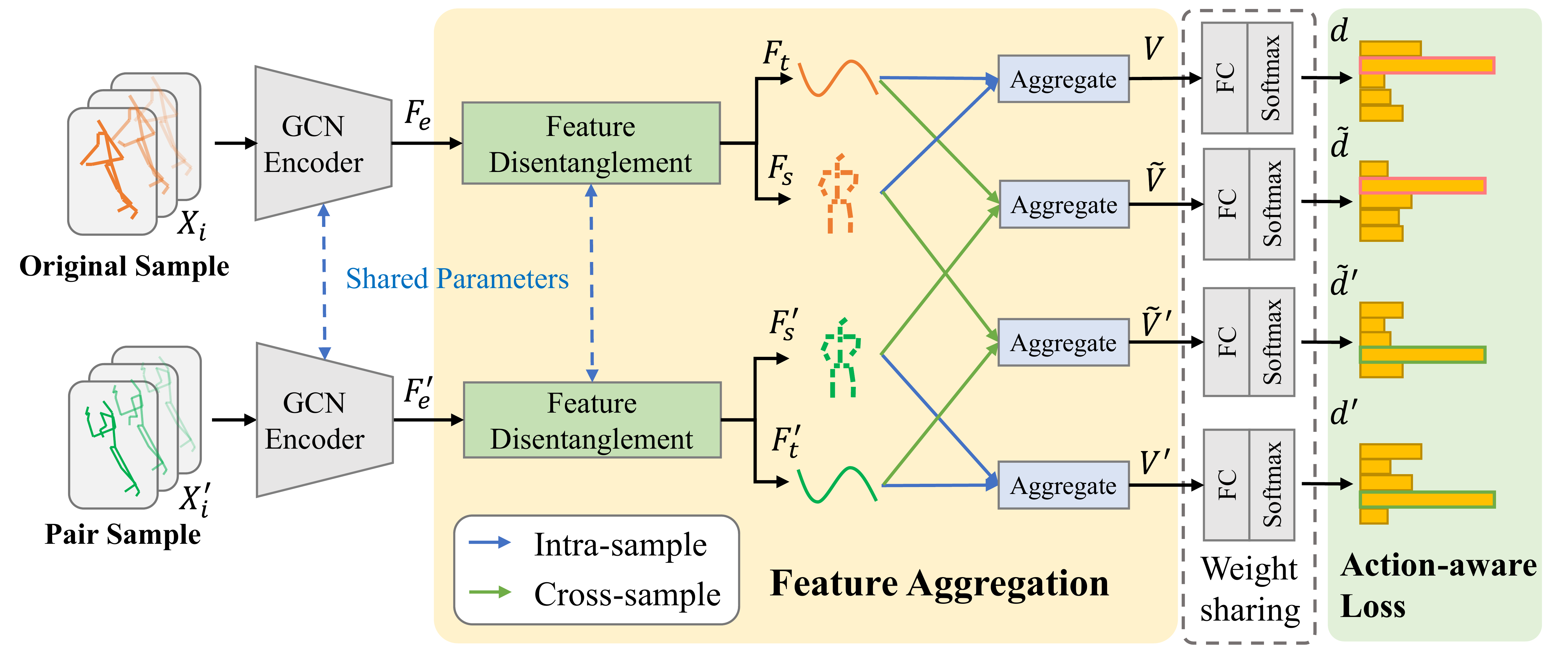}
  \caption{Overview of SkeletonX. In the training phase, an original sample ($X$) and its pair sample ($X^{'}$) are first encoded using a GCN backbone to features $F_e$ and $F_e'$, followed by disentanglement into spatial features ($F_s,F_s^{'}$) and temporal features ($F_t$,$F_t^{'}$). These features are processed through a feature aggregation module, where $V$ and $V'$ result from intra-sample aggregation, and $\widetilde{V}$ and $\widetilde{V}'$ from cross-sample aggregation. The aggregated features are then passed through a fully connected layer and softmax function to produce the predicted probability distribution $d$, $\widetilde{d}$, $\widetilde{d}'$, and $d'$. The model is finally optimized with our proposed action-aware loss.}
\label{fig:overview}
\end{figure*}

\section{Related Work}

\subsection{Skeleton-based Action Recognition}

Action recognition using skeleton data has attracted increasing attention for its inherent robustness and compact nature.
Early studies~\cite{duHierarchicalRecurrentNeural2015,zhangGeometricFeaturesSkeletonBased2017,zhangFusingGeometricFeatures2018,liuSkeletonBasedHumanAction2018} framed the task as a sequential modeling problem, employing Recurrent Neural Networks (RNNs) to solve it. However, RNNs suffer from gradient vanishing and long training time, which hinder the model from achieving better performance.

As human skeletons are naturally graph-structured data, Graph Convolutional Networks (GCNs) have gained significant attention in skeleton-based action recognition. Yan \etal\cite{yanSpatialTemporalGraph2018} pioneered a GCN-based approach for skeleton-based action recognition, incorporating partitioning strategies for skeletons and integrating spatial and temporal convolution to model skeleton sequences effectively. Given the significance of topology information in skeleton sequences, many scholars have focused on topology modeling through dynamic topology modeling\cite{shiTwoStreamAdaptiveGraph2019, liActionalStructuralGraphConvolutional2019, korbanDDGCNDynamicDirected2020, chenChannelwiseTopologyRefinement2021,tian2023skeletonbased, zhouBlockGCNRedefineTopology2024}, hierarchical modeling~\cite{leeHierarchicallyDecomposedGraph2023} and spatial-temporal channel aggregation\cite{chengDecouplingGCNDropGraph2020, liuDisentanglingUnifyingGraph2020, wangSkeletonbasedActionRecognition2022,tian2023skeletonbased, zhangCrossScaleSpatiotemporalRefinement2024}. Although previous GCN models are inherently capable of processing spatial and temporal information, we propose enhancing their performance through a cross-sample aggregation training paradigm.

\subsection{One-shot Skeleton-based Action Recognition}
\label{sec:related_LowShotSAR}
Motivated by the human ability to recognize new concepts with one example, one-shot skeleton action recognition has attracted increasing attention in recent years. Liu \etal~\cite{liuNTURGB1202020} presented a large-scale dataset with one-shot skeleton-based action recognition benchmarks, and introduced an APSR framework to emphasize the body parts relevant to the novel actions as a solution. Sabater \etal~\cite{sabaterOneShotActionRecognition2021} proposed a motion encoder based on geometric information, which is robust to various movement kinematics. Wang \etal~\cite{wangTemporalViewpointTransportationPlan2022} introduced JEANIE, a method that aligns the joint of temporal blocks and simulates viewpoints during meta-learning. Memmesheimer \etal~\cite{memmesheimerSkeletonDMLDeepMetric2022} introduced an image-based skeleton representation and trained an embedder to project the images into an embedding vector. Zhu \etal~\cite{zhuAdaptiveLocalComponentawareGraph2023} proposed a novel matrix-based distance metric to determine the spatial and temporal similarity between two skeleton sequences. Peng\etal~\cite{peng2023delving} studied occluded skeleton recognition scenarios and proposed Trans4SOAR to fusion diverse inputs. Yang \etal~\cite{yangOneShotActionRecognition2024} proposed a multi-spatial and multi-temporal skeleton feature representation, and a matching technique that handles scale-wise and cross-scale features.

Previous research mainly focuses on matching strategies, distance metrics, or improving model structure to achieve better feature representation. We propose that the inherent correlations among the limited labeled samples hold rich information for enhancing representation learning. Additionally, these methods are designed under meta-learning paradigms, limiting the applications in other cases. Distinct from previous methods, we introduce a novel approach involving a sample pair selection strategy and aggregation method to fuse pair samples. Our method can be further expanded to more scenarios while maintaining compatibility with established methods.

\subsection{Limited-scale settings}

Some work has conducted experiments on randomly resampled data with few-shot~\cite{maLearningSpatialPreservedSkeleton2022, liuParallelAttentionInteraction2023} or fully-supervised~\cite{duan2022dgstgcn} settings. However, we note that random sampling is not realistic for applications. The full dataset includes diverse action performers and camera setups. Several samples are likely obtained from nearly every performer and setup through random sampling, while each performer or setup contributes only a few samples. When training samples are collected for specific applications, it would be more practical to recruit a few volunteers with some camera setups and collect more samples. Therefore, we propose a new simulation of the scenarios with a new limited-scale sampling strategy.

\section{Method}

In this section, we first formulate the problem settings. An overview of our method is provided in Fig.~\ref{fig:overview}. The remainder of this section introduces the sample pairs selection strategy, cross-sample feature aggregation, and action-aware loss function within our proposed pipeline.

\begin{figure}[t]
\centering
\includegraphics[width=\linewidth]{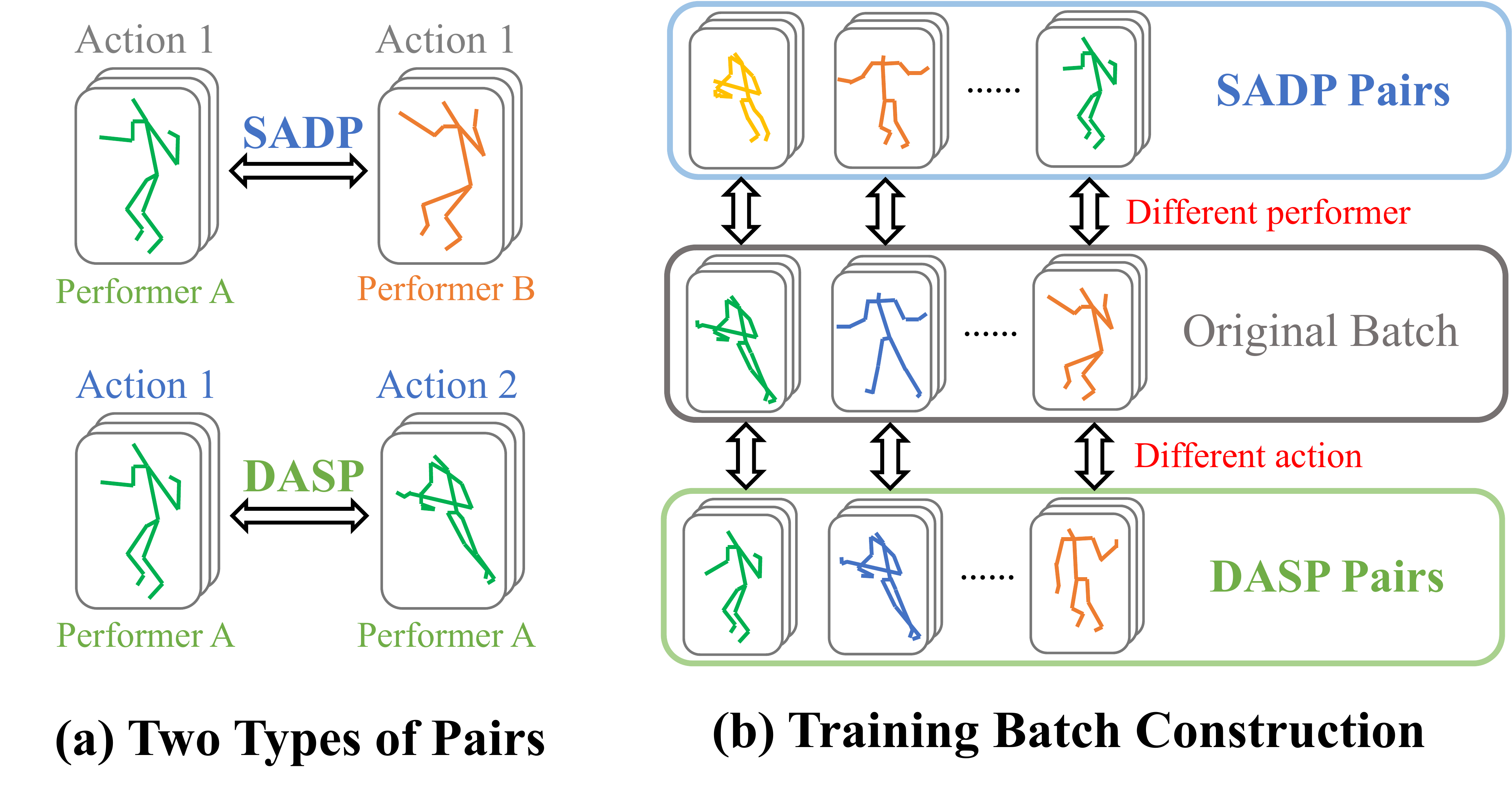}
\caption{Illustration for (a) two types of sample pairs: DASP (Different Action, Same Performer) and SADP (Same Action, Different Performer). (b) Training-time batch construction. For each sample within the original batch, its DASP sample and SADP sample form two batches, which are used in the aggregation step of our proposed pipeline.}
\label{fig:sample_selection}
\end{figure}

\subsection{Preliminary and Problem Definition}
\label{sec:prelim}

\textbf{Skeleton-based Action Sequence.} A human skeleton can be defined with a set of vertices $S=(x_1,x_2,\ldots,x_V)$, where $V$ is the number of joints provided by the dataset, and each vertex $x_i$ is denoted with 3-dimensional coordinate. One action sequence $X_i=(S_1,S_2,\ldots,S_T)$ consists of $T$ frames of human skeleton sequence, therefore $X_i\in\mathbb{R}^{T\times V\times 3}$. The label of the skeleton sequence is denoted as one-hot label $y\in\mathbb{R}^K$, where $K$ is the number of action categories.

\textbf{One-shot Action Recognition} aims to identify the novel action with only one exemplar per category. Training set is denoted as $D_{b}$, which contains $B$ base classes $C_{b}=\{c_1,c_2,\ldots,c_B\}$. Testing set contains $N$ novel classes $C_{n}=\{c_1,c_2,\ldots,c_N\}$. The training set shares no common class with the testing set, \ie~$C_{b}\cap C_{n}=\emptyset$. The models are first trained with $C_b$, and act as feature extractors during inference. For most of the one-shot methods, the final prediction is made by measuring the distances between test samples and exemplars in feature space.

\textbf{Limited-scale Setting} aims to train the network from scratch with a limited number of samples collected from the target scenario in a fully supervised manner. When dealing with varying numbers of joints or significant domain gaps, the existing one-shot learning paradigm may not directly apply to real-world scenarios. We propose the limited-scale setting, which collects a handful of samples for each action category. Formally, the limited-scale dataset is denoted as $D_{LS} = \{X_1^1,\ldots,X_1^N,X_2^1,\ldots,X_2^N,\ldots,X_C^1,\ldots,X_C^N\}$, where $X_i^j$ is the $j$-th skeleton sequence drawn from action class $i$. $N$ is the number of samples drawn for each class, and $C$ represents the number of classes.

In contrast to prior work~\cite{maLearningSpatialPreservedSkeleton2022, duan2022dgstgcn, liuParallelAttentionInteraction2023} which randomly selects samples from the training set, this setting controls the number of subjects and camera perspectives to minimize costs. To simulate this challenging scenario, we resample the training data while keeping the testing data fixed in cross-subject fully supervised learning. Further details are provided in Section~\ref{sec:eval_proto}.

\subsection{Sample Pairs Selection Strategy}

We define two types of sample pairs based on two key attributes of skeleton-based action recognition: performer and action, as illustrated in Fig.~\ref{fig:sample_selection}(a). For example, consider a sample that depicts Action 1, performed by Performer A. The corresponding DASP partner would be another skeleton sequence representing action 2, also performed by Performer A. Conversely, the SADP partner would be the same action category but performed by a different individual, Performer B. By fixing one attribute, the model can focus on the other during the subsequent cross-sample aggregation process.

During each training step, a batch is constructed by randomly selecting skeleton sequences, as illustrated in Fig.~\ref{fig:sample_selection}(b). For every sequence within the batch, one sample from each pair type is drawn to form two batches for aggregation. The strategy enables us to disentangle and aggregate samples. It is worth noting that this procedure can be performed dynamically by a modified data loader during training, or statically through data pre-processing to reduce the overhead of sample selection.

\begin{figure}
    \centering
    \includegraphics[width=\linewidth]{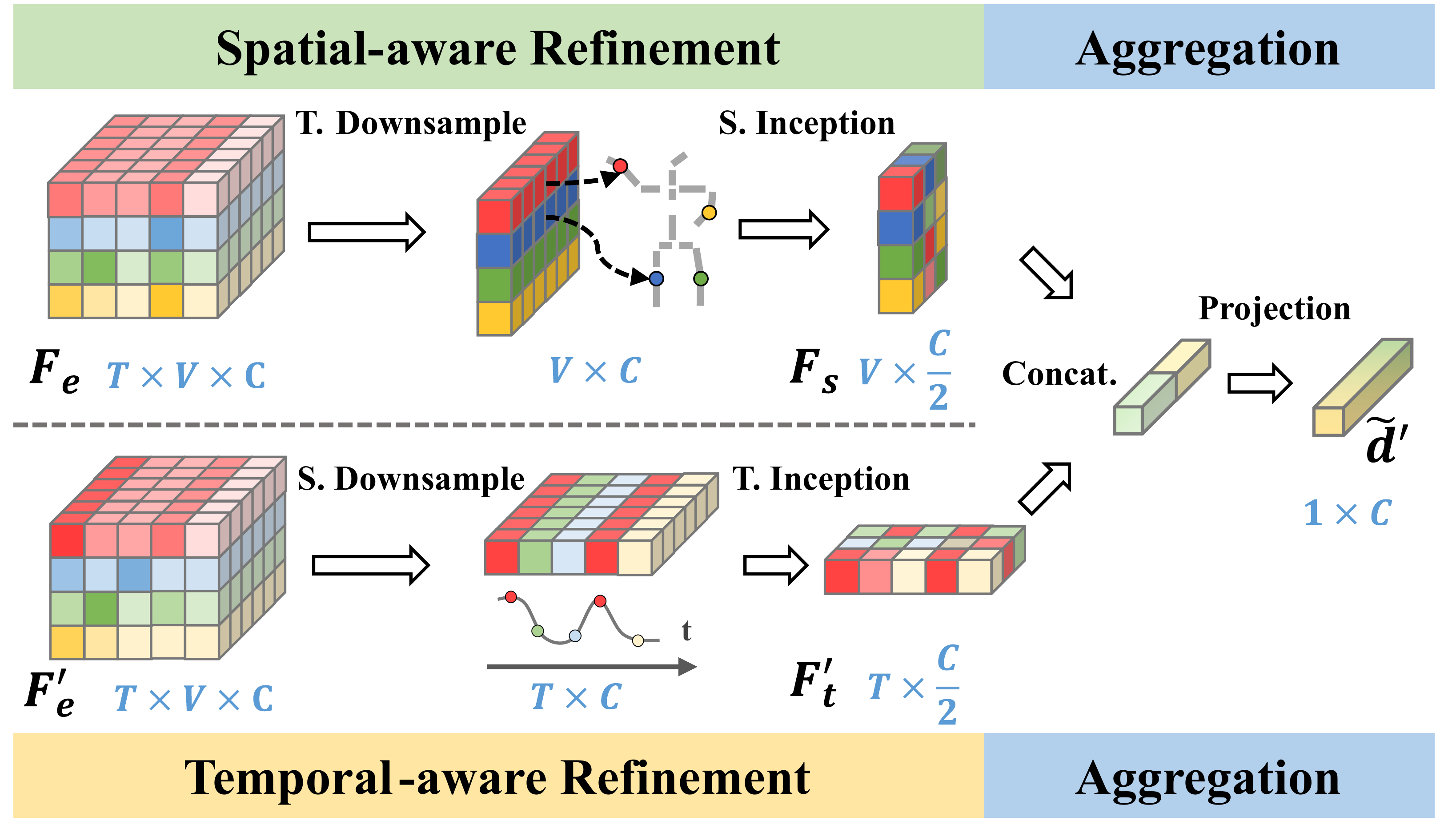}
    \caption{Illustration for cross-sample aggregation. Features of the original sample $F_e$ and its pair sample $F_e'$ are refined by two branches respectively and then aggregated to produce the final prediction $\widetilde{d}'$.}
    \label{fig:aggr_module}
\end{figure}

\subsection{Cross-sample Feature Aggregation}
\label{sec:feat_disent}
Most of the GCN-based methods~\cite{yanSpatialTemporalGraph2018,shiTwoStreamAdaptiveGraph2019,chenChannelwiseTopologyRefinement2021,wangSkeletonbasedActionRecognition2022} focus on designing GCN building blocks while adhering to the overall structure proposed by~\cite{yanSpatialTemporalGraph2018}. In the general structure, the skeleton sequence is encoded to features using a GCN encoder. These encoded features are then processed through a simple Global Average Pooling (GAP) layer, producing sequence-level feature vectors for final prediction. 

As previous methods have proposed sophisticated designs on spatial and temporal modeling within GCN encoders, the design philosophy of our aggregation module is to enhance the capabilities of existing modules instead of replacing them. Therefore, we propose a concise feature aggregation module on the output of the GCN encoders to replace the GAP layer. As shown in Fig.~\ref{fig:aggr_module}, it includes two refinement modules focusing on spatial and temporal information respectively, which provide fine-grained decoupled features for aggregation.

We take the cross-sample aggregation process of $\widetilde{V}\,^{'}$ shown in Fig.~\ref{fig:overview} as an example. Given the encoded feature $F_{e}\in\mathbb{R}^{T\times V\times C}$, the spatial-aware refinement is defined as:

\begin{equation}
  \label{eq:disent_spa}
  F_s = \mathbf{ReLU}(\mathbf{BN}(W^c_s\cdot \mathbf{AP}_t(F_{e}) + b^c_s))
\end{equation}

where $W^c_s$ and $b^c_s$ are channel-wise learnable weights that aggregate information of each joint, enabling the network to capture spatial patterns. $\mathbf{AP}_t$ indicate Average Pooling along the temporal axis, it down-samples the feature on the temporal axis, allowing the module to focus on spatial information of joints. $\mathbf{BN}$ represents batch normalization, and $\mathbf{ReLU}$ is the activation function.

Temporal-aware refinement branch uses $\mathbf{AP}_s$ along spatial axis and its own parameters $W^c_t$, $b^c_t$, defined as:

\begin{equation}
  \label{eq:disent_temp}
  F_t^{'} = \mathbf{ReLU}(\mathbf{BN}(W^c_t\cdot \mathbf{AP}_s(F_{e}^{'}) + b^c_t))
\end{equation}

By disentangling the feature, the spatial feature carries the performer's properties (i.e., performer-related), while the temporal feature captures the action momentum along the timeline (i.e., action-related). As the gradients flow forward, the module also encourages the GCN encoder to distinguish the two aspects of actions better.

Aggregation of two decoupled features is done through concatenation and projection layer, defined as:

\begin{equation}
    \label{eq:aggr_func}
    \widetilde{d}\,^{'} = W_p \cdot\mathbf{Concat}(\mathbf{AP}_s(F_s), \mathbf{AP}_t(F_t^{'})) + b_p
\end{equation}

where the average pooling (AP) summarizes the performer-related and action-related feature, and a learnable projection layer with parameters $W_p$ and $b_p$ to further fuse the feature. $\mathbf{Concat}$ indicates concatenation along the feature dimension. We discover that concatenation is surprisingly effective by the ablation studies. This effectiveness may be attributed to our training strategy, which guides the sophisticated designed modules in GCN encoders to focus on different aspects, thus enhancing performance.

\subsection{Loss Function}
\label{sec:objective}
Based on the design of our proposed module, we present an action-aware loss function to encourage the model to learn different representations related to actions among various subjects. As shown in Fig.~\ref{fig:overview}, the aggregated features are fed into a fully connected (FC) layer followed by a SoftMax layer to produce the final predictions. After feature aggregation, two source samples produce predictions based on four aggregated features. We employ Cross-Entropy ($\mathbf{CE}$) as the loss 
function based on action-aware labels to train the network, where the label of the aggregated feature is determined by the source of the action-related feature. 

Based on two aggregated samples, we present two types of loss functions: inter-sample loss and intra-sample loss.

$\widetilde{V}$ has the same label as original sample $y$, while $\widetilde{V}^{'}$ is labeled according to the paired sample $y^{'}$. The loss is defined as Equation~\ref{eq:DASP_loss}, where $\widetilde{d}$ and $\widetilde{d}\,^{'}$ represent the probabilities of the two aggregated samples illustrated in Fig.~\ref{fig:overview}.

\begin{equation}
  \label{eq:DASP_loss}
  L_{DASP} = \mathbf{CE}(\widetilde{d}, y) + \mathbf{CE}(\widetilde{d}\,^{'}, y^{'})
\end{equation}

The aggregation is similarly performed for the SADP pair. Since ``SA'' stands for ``Same Action'', the paired sample label $y^{'}$ is the same as the original sample $y$. The SADP loss is formulated as:
\begin{equation}
  \label{eq:SADP_loss}
  L_{SADP} = \mathbf{CE}(\widetilde{d}, y) + \mathbf{CE}(\widetilde{d}\,^{'}, y)
\end{equation}

The features of the original sample and two paired samples are also aggregated, as indicated by the blue arrows in Fig.~\ref{fig:overview}. The loss of intra-sample aggregation is formulated as:
\begin{equation}
  \label{eq:intra_loss}
  L_{Intra} = \mathbf{CE}(d, y) + \mathbf{CE}(d^{'}, y) + \mathbf{CE}(d^{'}_{*}, y^{'})
\end{equation}
where $d^{'}$ indicates the SADP pair, and $d^{'}_*$ indicates the DASP pair.

Finally, CE losses from different training objectives are combined to form the final learning objective function:
\begin{equation}
  \label{eq:tot_loss}
  L=L_{Intra} + w_x\left( L_{DASP} + L_{SADP}\right)
\end{equation}

where $w_x$ is the balance hyper-parameter for cross-sample aggregation.

\subsection{Theoretical Analysis}
\label{sec:theory}
In this section, we analyze the feature aggregation technique through Information Bottleneck (IB) theory~\cite{tishby2000information,alemi2017deep}, and propose that it provides more information for the model under limited training data. Previous research on skeleton-based action recognition~\cite{chi2022infogcn} has demonstrated the correlation between mutual information and improved performance. Unlike the prior approach, which focused on designing the network structure, we enhance mutual information via sample aggregation, thus boosting performance. Additionally, we validate our theory through experiments in Section~\ref{sec:MI_estimation}.

In the following analysis, we focus on cross-sample aggregation for DASP. Nevertheless, the subsequent formulation also applies to the alternative case. As described in the previous section, the skeleton sequence $X_i$ and its paired sample $X_i'$ are first encoded by encoder $f_\theta$ to features $F_i$ and $F_i'$. To simplify the notation, we omit subscripts and superscripts in the formulas, representing the samples as $X$ and $X'$, and the corresponding features as $F$ and $F'$.

\begin{equation}
  \label{eq:encoding}
  F=f_\theta(X) \qquad F'=f_\theta(X')
\end{equation}

The aggregation module $A$ takes two features as input, with the first feature encoding action information and the second encoding performer information. The aggregated features are denoted as $\widetilde{V}$ and $\widetilde{V}'$ following Fig.~\ref{fig:overview}.

\begin{equation}
  \label{eq:aggr}
  \widetilde{V}=A(F, F') \qquad \widetilde{V}'=A(F', F)
\end{equation}

We analyze the first aggregated feature $\widetilde{V}$, denoted as $V$ for simplicity. The label of $X$ is denoted as $Y$. Following previous work~\cite{tishby2000information, alemi2017deep}, the information bottleneck objective function can be formulated as

\begin{equation}
  \label{eq:ib}
  R_{IB}=I(D;Y)-\beta I(D;X)
\end{equation}

where $I(\cdot;\cdot)$ denotes the mutual information between two random variables, $\beta$ is the Lagrange multiplier, and $D$ represents the intermediate feature. According to IB theory, the model improves by learning a concise representation while retains good predictive capability, achieved by \textbf{maximizing} the objective function $R_{IB}$.

In our work, the aggregated feature $V$ replaces the feature $D$ of the baseline method through sample aggregation. Given the data processing inequality, 

\begin{equation}
  \label{ieq:data_processing}
  I(V;Y)\leq I(F,F';Y)
\end{equation}

where $I(F,F';Y)$ represents the mutual information between features of the pair samples and the corresponding label $Y$. If the aggregation method encodes the information efficiently, $I(V;Y)$ will approach its upper bound $I(F,F';Y)$. The mutual information of $V$ and $Y$ can be expressed as

\begin{align}
I(V;Y) &= E_{V,Y}[\mathbf{log}\frac{p(Y|V)}{p(Y)}] \\
&= E_{V,Y}[\mathbf{log}(p(Y|V))]+H(Y)
\end{align}

where $E$ denotes the mathematical expectation, $p$ denotes the probability distribution, and $q$ is the variational estimate of $p$. $H$ indicates the entropy function. According to~\cite{alemi2017deep},

\begin{align}
I(V;Y)&= E_{V,Y}[\mathbf{log}(p(Y|V))]+H(Y)\\
&\geq E_{V,Y}[\mathbf{log}(q(Y|V))]+H(Y) \label{eq:ieq_MI}
\end{align}

From equations (\ref{eq:ib}) and (\ref{eq:ieq_MI}), we obtain

\begin{equation}
  \label{eq:ieq_all}
  I(F,F';Y) \geq I(V;Y) \geq E_{V,Y}[\mathbf{log}(q(Y|V))]+H(Y)
\end{equation}

In our action-aware loss, the model is optimized using cross-entropy loss, which is equivalent to

\begin{equation}
  \label{eq:ce_loss}
  \mathbf{Maximize}~~ E_{V,Y}[\mathbf{log}(q(Y|V))]
\end{equation}

By maximizing the lower bound of mutual information $I(V;Y)$, we optimize the first term of the information bottleneck theory, which improves the predictive capability of the model.

\begin{table}[tb]
    \centering
    \renewcommand{\arraystretch}{1.2}
    \caption{Ablation on major components in \textbf{limited-scale setting} and \textbf{one-shot setting} (\textit{O.S.}). Top-1 accuracy (\%) is displayed. \textit{Pair} indicates sample pairs selection, \textit{Disent.} indicates feature disentanglement, \textit{X-Aggr.} indicates cross-sample aggregation.}
    \begin{tabularx}{\columnwidth}{l|>{\centering\arraybackslash}X>{\centering\arraybackslash}X>{\centering\arraybackslash}X|>{\centering\arraybackslash}X}
      \hline
      Method & NTU & NTU120 & PKU & NTU120(O.S.)\\
      \hline
      Baseline & 41.5 & 19.7 & 28.5 & 39.0\\
      ~~+ Pair & 46.4 & 23.6 & 31.4 & 43.1\\
      ~~+ Disent. & 47.2 & 24.4 & 36.0 & 47.4\\
      ~~+ X-Aggr. & \textbf{52.3} & \textbf{27.0} & \textbf{37.0} & \textbf{48.2}\\
    \hline
    \end{tabularx}
    \label{tab:abl_limit}
  \end{table}

\begin{table}[tb]
    \centering
    \renewcommand{\arraystretch}{1.2}
    \caption{Ablation on balance hyper-parameter in NTU RGB+D \textbf{limited-scale setting}. Results are shown in Top-1 Accuracy (\%).}
    \begin{tabularx}{\columnwidth}{>{\centering\arraybackslash}X|>{\centering\arraybackslash}X>{\centering\arraybackslash}X>{\centering\arraybackslash}X>{\centering\arraybackslash}X}
    \hline
    \multirow{2}{*}{$w_x$} & \multicolumn{4}{c}{Samples per class}\\
    & 10 & 20 & 30 & 50\\
    \hline
    \textbf{0.1} & \textbf{52.3} & \textbf{62.6} & \textbf{67.9} & \textbf{73.4}\\
    0.2 & 51.4 & 60.7 & 67.3 & 72.7\\
    0.5 & 52.1 & 62.0 & 67.5 & 73.1\\
    1.0 & 51.2 & 61.9 & 67.6 & 72.5\\
    \hline
    \end{tabularx}
    \label{tab:abl_hyper}
\end{table}

\section{Experiment}
\label{sec:exp}

\subsection{Datasets}
We select three widely adopted datasets that provide subject (performer) attributes of each action sequence to implement and evaluate our proposed method.

\textbf{NTU-RGB+D (NTU)}~\cite{shahroudyNTURGBLarge2016} dataset consists of 56,880 video samples captured using Microsoft Kinect v2, which is one of the most widely used dataset in skeleton-based action recognition. Depth cameras with 3 different horizontal angle settings provide precise 3D coordinates of joints. Actions in this dataset are performed by 40 human subjects and are categorized into 60 classes. The dataset provides two benchmarks: Cross Subject (CSub) and Cross View (CView).~\textit{CSub} setting divides the dataset according to the subjects. The training set includes 20 subjects, and the testing set includes the other 20 subjects.~\textit{CView} setting splits the dataset by the camera views, while the subjects are the same in the training and testing phase.

\textbf{NTU-RGB+D 120 (NTU-120)}~\cite{liuNTURGB1202020} dataset extends NTU-RGB+D~\cite{shahroudyNTURGBLarge2016} to 113,945 human skeleton sequences to 120 action categories performed by 106 subjects, and 32 different views that vary in camera setup and background.
It also provides an additional benchmark: One-shot learning, where 20 classes are assigned as novel categories with one exemplar sample per class.

\textbf{PKU-MMD}~\cite{liuBenchmarkDatasetComparison2020} dataset is a large-scale multi-modality benchmark for 3D skeleton action recognition. It contains around 28,000 action instances and comprises two subsets under different settings: Part I and Part II. We employ part II in this paper, primarily because it provides sequence-wise subject label. Part II contains around 6,900 action instances with 41 action classes, performed by 13 subjects. Part II is also more challenging than Part I due to short action intervals, concurrent actions, and heavy occlusion.

\begin{table}[tb]
  \centering
  \renewcommand{\arraystretch}{1.2}
  \caption{Ablation study on feature disentanglement by setting part of the disentangled feature to zero in \textbf{limited-scale setting} (10 samples per class) on NTU-RGB+D dataset. Top-1 accuracy (\%) is shown in the table. \textit{w/o Action Feat.} indicates mask action feature, and \textit{w/o Performer Feat.} indicates masking the performer feature.}
  \begin{tabularx}{\columnwidth}{@{}l>{\centering\arraybackslash}X>{\centering\arraybackslash}X>{\centering\arraybackslash}X@{}}
    \hline
    Method & CTR-GCN & ST-GCN & TCA-GCN\\
    \hline
    Baseline & 41.5 & 35.4 & 42.1\\
    \hline
    Ours & \textbf{52.3} & \textbf{47.9} & \textbf{50.9}\\
    ~~w/o Action Feat. & 11.5 & 11.2 & 13.0\\
    ~~w/o Performer Feat. & 49.9 & 36.0 & 48.1\\
  \hline
  \end{tabularx}
  \label{tab:abl_decouple}
\end{table}

\begin{table}[tb]
  \centering
  \renewcommand{\arraystretch}{1.3}
  \caption{Accuracy (\%) comparison on \textbf{one-shot setting} with different numbers of training classes on NTU-RGB+D 120 dataset. The best results are shown in \textbf{bold}, and the second best results are shown with \underline{underline}.}
  \begin{tabularx}{\columnwidth}{@{}l>{\centering\arraybackslash}X>{\centering\arraybackslash}X>{\centering\arraybackslash}X>{\centering\arraybackslash}X>{\centering\arraybackslash}X@{}}
    \hline
    Training Classes & 20 & 40 & 60 & 80 & 100\\
    \hline
    APSR~\cite{liuNTURGB1202020} & 29.1 & 34.8 & 39.2 & 42.8 & 45.3\\
    SL-DML~\cite{memmesheimerSLDMLSignalLevel2021} & 36.7 & 42.4 & 49.0 & 46.4 & 50.9\\
    Skeleton-DML~\cite{memmesheimerSkeletonDMLDeepMetric2022} & 28.6 & 37.5 & 48.6 & 48.0 & 54.2\\
    uDTW~\cite{wangUncertaintyDTWTimeSeries2022} & 32.2 & 39.0 & 41.2 & 45.3 & 49.0\\
    JEANIE~\cite{wangTemporalViewpointTransportationPlan2022} & 38.5 & 44.1 & 50.3 & 51.2 & 57.0\\
    SMAM~\cite{liSMAMSelfMutual2023} & 35.8 & 46.2 & 51.7 & 52.2 & 56.4\\
    ALCA-GCN~\cite{zhuAdaptiveLocalComponentawareGraph2023} & 38.7 & 46.6 & 51.0 & 53.7 & 57.6\\
    STA-MLN~\cite{liSpatialTemporalAdaptiveMetric2024} & 42.5 & 48.8 & 53.1 & 54.3 & 59.9\\
    M\&C-scale\cite{yangOneShotActionRecognition2024} & \underline{44.1} & \textbf{55.3} & \underline{60.3} & \underline{64.2} & \underline{68.7}\\
    \hline
    SkeletonX(Ours) & \textbf{48.2} & \underline{54.9} & \textbf{61.6} & \textbf{65.6} & \textbf{69.1}\\
    
  \hline
  \end{tabularx}
  \label{tab:os_cls}
\end{table}

\subsection{Evaluation Protocols}
\label{sec:eval_proto}
To validate the data efficiency of our proposed method, we conduct experiments in two settings: the one-shot setting and the limited-scale setting.

\textbf{One-shot Learning}. We follow the one-shot evaluation protocol proposed in~\cite{liuNTURGB1202020}, which selects 100 base classes and 20 novel classes with one fixed exemplar. Samples from base classes are used for training, and samples other than the exemplars in the novel classes are used for testing. For NTU RGB+D and PKU-MMD datasets, we follow previous work~\cite{peng2023delving, yanCrossGLGLLMGuides2024a, yangOneShotActionRecognition2024} to assign 10 novel classes with one fixed exemplar.

\textbf{Limited-Scale Training}. We sample from the original cross-subject training set to simulate datasets with different scales and explore the potential of performance improvement with continuous data collection.
Specifically, we draw 10, 20, 30, and 50 samples per class for the training set. Instead of randomly drawn samples from the training set~\cite{maLearningSpatialPreservedSkeleton2022}, we limit the number of subjects and setups to reduce the cost of finding volunteers and setting up different camera views. As illustrated in Alg.~\ref{alg:sample_selection}, we select $P$ performers with the least setups when sampling from each action category based on cross-subject setting. Specifically, we first create a table $T$ with columns `performer', `setup', and `skeleton\_filename'. Each row corresponds to a sample in the training set, initially sorted alphabetically by `skeleton\_filename'. The testing set is kept intact to ensure adequate evaluation.

\begin{algorithm}[tb]
\caption{Limited-scale Sample Selection Algorithm}
\label{alg:algorithm}
\textbf{Input}: Table \( T \) with columns: performer ID, setup, and action ID\\
\textbf{Parameter}: Number of samples per class \( N \), Number of action categories \( C \), Number of performer \( P \)\\
\textbf{Output}: Trimmed Table \( T' \)
\begin{algorithmic}[1] 
\STATE Drop rows with performer ID larger than $P$
\STATE Sort $T$ by setup (primary) and performer ID (secondary)
\STATE Let $c=1$
\WHILE{$c\le C$}
\STATE Select the first $N$ rows in $T$ for class $c$, and append them to \( T' \).
\ENDWHILE
\STATE \textbf{return} \( T' \)
\end{algorithmic}
\label{alg:sample_selection}
\end{algorithm}

\subsection{Implementation Details}
\label{sec:exp_implem}

Experiments are conducted using two Nvidia RTX 2080 Ti GPUs for the CTR-GCN~\cite{chenChannelwiseTopologyRefinement2021}, ST-GCN~\cite{yanSpatialTemporalGraph2018} and BlockGCN~\cite{zhouBlockGCNRedefineTopology2024} encoders. Two Nvidia V100 GPUs are used for experiments on TCA-GCN~\cite{wangSkeletonbasedActionRecognition2022} and SkateFormer~\cite{do2024skateformer} due to their excessive parameters. We follow the data pre-processing procedures described in~\cite{chenChannelwiseTopologyRefinement2021}, which remove empty frames and resize each video clip to 64 frames. For all GCN-based backbones, we apply the SGD optimizer with a momentum of 0.9 and a weight decay of 0.0004 to train the model. For the transformer-based model SkateFormer, we follow the training protocol as the original paper~\cite{do2024skateformer}. The balance parameter $w_x=0.1$ is set for two settings based on the ablation studies. All experiments are conducted in the joint modality.

During the inference phase, the feature disentanglement module and aggregate module are retained with the trained parameters and only perform intra-sample aggregation. The output of the SoftMax layer indicates the probability of each action category.

\textbf{Limited-Scale Training.} The model is trained for 65 epochs with an initial learning rate of 0.1. The learning rate decays by a factor of 0.1 at epoch 35 and 55. To stabilize the training process, a warm-up strategy is employed during the first 5 epochs. Following~\cite{chenChannelwiseTopologyRefinement2021}, the batch size is set at 64 for NTU RGB+D, NTU RGB+D 120 and PKU-MMD in baseline methods. To accommodate the additional samples introduced by our method, the batch size is reduced to 32 when training with our approach.

\textbf{One-shot Learning.} We adopt~\cite{chenChannelwiseTopologyRefinement2021} as the GCN encoder for one-shot learning. The model is trained for 25 epochs for better performance. The initial learning rate is set at 0.1 with a warm-up in the first 5 epochs. The learning rate decays by 0.1 at epoch 10 and 15. The model is trained on base categories in a fully-supervised setting. During inference, the FC layer and SoftMax layer are removed to generate features for action sequences. Protonet~\cite{snellPrototypicalNetworksFewshot2017} is used as the metric learning method for feature matching.

\begin{table*}[tb]
  \centering
  \renewcommand{\arraystretch}{1.2}
  \caption{Ablation on feature aggregation methods in \textbf{limited-scale setting}. Top-1 accuracy (\%) is displayed.}
  \begin{tabularx}{\textwidth}{l|*{5}{>{\centering\arraybackslash}X}|*{5}{>{\centering\arraybackslash}X}}
    \hline
    \multirow{2}{*}{Method} & \multicolumn{5}{c|}{NTU RGB+D 120} & \multicolumn{5}{c}{NTU RGB+D}\\
     & 10 & 20 & 30 & 50 & Average & 10 & 20 & 30 & 50 & Average\\
    \hline
    Concat & \textbf{27.0} & \textbf{32.6} & \textbf{38.1} & \textbf{43.5} & \textbf{35.3} & 52.3 & 62.6 & \textbf{67.9} & 73.4 & \textbf{64.1}\\
    MatMul & 24.3 & 31.4 & 36.2 & 41.1 & 33.3 & \textbf{52.8} & 61.0 & 67.0 & 72.4 & 63.3\\
    Cross-attn & 26.0 & 30.2 & 36.4 & 43.0 & 33.9 & 52.6 & \textbf{62.7} & 67.0 & \textbf{73.5} & 64.0\\
  \hline
  \end{tabularx}
  \label{tab:abl_feat_aggr}
\end{table*}

\begin{table*}[tb]
  \centering
  \renewcommand{\arraystretch}{1.2}
  \caption{Accuracy (\%) in the \textbf{one-shot setting}.~\textit{NTU} and \textit{NTU-120} refer to NTU-RGB+D and NTU-120 respectively.~\textit{Param.} indicates number of parameters. FLOPs of the backbone model are presented in parentheses~(if applicable). The best results are highlighted in~\textbf{bold}.}
  \begin{threeparttable}
  \begin{tabularx}{\textwidth}{@{}l*{5}{>{\centering\arraybackslash}X}@{}}
    \toprule
    Method & NTU & NTU120 & PKU-MMD(II) & Param.(M) & GFLOPs\\
    \midrule
    \textbf{Meta-learning Methods} \\
    ProtoNet~\cite{snellPrototypicalNetworksFewshot2017} & 75.0 & 66.0 & 37.0 & 1.46 & 1.79~(1.79)\\
    FEAT~\cite{yeFewShotLearningEmbedding2020} & 80.5 & 65.1 & 32.3 & 1.70 & 1.80~(1.79) \\
    \midrule
    \textbf{Other Methods}\\
    CATFormer\cite{longSTEPCATFormerSpatialTemporal2023} & 73.2 & 57.2 & 31.8 & 4.17 & 18.35\\
    FR-Head\cite{zhouLearningDiscriminativeRepresentations2023a} & 78.1 & 65.4 & 34.3 & 1.99 & 1.79\\
    \midrule
    \textbf{Matching-based Methods}\\
    SL-DML~\cite{memmesheimerSLDMLSignalLevel2021} & - & 50.9 & - & 11.2\tnote{\dag} & 23.8\tnote{\dag}\\
    Skeleton-DML~\cite{memmesheimerSkeletonDMLDeepMetric2022} & - & 54.2 & - & 11.2\tnote{\dag} & 23.8\tnote{\dag}\\
    SL-DML (CTR-GCN~\cite{chenChannelwiseTopologyRefinement2021}) & - & 43.9\tnote{\dag} & - & 1.60\tnote{\dag}~(1.46) & 9.2\tnote{\dag}~(1.79)\\
    Trans4SOAR (Small)~\cite{peng2023delving} & - & 56.3 & - & 23.1 & 34.1\\
    Trans4SOAR (Base)~\cite{peng2023delving} & - & 57.1 & - & 43.8 & 47.9\\
    Koopman~\cite{wangNeuralKoopmanPooling2023} & - & 68.1 & - & 1.46 & 1.96~(1.79)\\
    S-scale\cite{yangOneShotActionRecognition2024} & 77.4 & 63.2 & - & 3.78 & 30.4~(3.99)\\
    M-scale\cite{yangOneShotActionRecognition2024} & 81.6 & 67.6 & - & 15.12 & 79.6~(12.85)\\
    M\&C-scale\cite{yangOneShotActionRecognition2024} & 82.7 & 68.7 & - & 15.12 & 79.9~(12.85)\\
    \midrule
    Ours & \textbf{83.2} & \textbf{69.1} & \textbf{38.3} & 1.53~(1.46) & 1.80~(1.79)\\
  \bottomrule
  \end{tabularx}
    \begin{tablenotes}
    \footnotesize          
    \item[\dag] Reported in paper~\cite{peng2023delving}.
    \end{tablenotes}
    \end{threeparttable}
  \label{tab:os_flops}
\end{table*}

\begin{table*}[tb]
  \centering
  \renewcommand{\arraystretch}{1.2}
  \caption{Top-1 accuracy (\%) when combining our proposed method to multiple skeleton-based action recognition GCN backbones in \textbf{limited-scale setting}. The best top-1 accuracy is in \textbf{bold}. \textit{R.R.} indicates Random Rotation.}
  \begin{tabularx}{\textwidth}{l|>{\centering\arraybackslash}X>{\centering\arraybackslash}X>{\centering\arraybackslash}X>{\centering\arraybackslash}X|>{\centering\arraybackslash}X>{\centering\arraybackslash}X>{\centering\arraybackslash}X>{\centering\arraybackslash}X|>{\centering\arraybackslash}X>{\centering\arraybackslash}X>{\centering\arraybackslash}X>{\centering\arraybackslash}X}
    \hline
    Methods & \multicolumn{4}{c|}{PKU-MMD (II)} & \multicolumn{4}{c|}{NTU} & \multicolumn{4}{c}{NTU-120}\\
     & 10 & 20 & 30 & 50 & 10 & 20 & 30 & 50 & 10 & 20 & 30 & 50\\
    \hline
    \hline
    ST-GCN~\cite{yanSpatialTemporalGraph2018} & 22.3 & 34.8 & 43.5 & 51.6 & 34.9 & 51.7 & 57.4 & 66.0 & 18.9 & 21.2 & 25.7 & 35.4\\
    w/ Mixup & 25.9 & 36.3 & 42.6 & 51.9 & 37.3 & 53.1 & 59.7 & 67.0 & 19.4 & 23.1 & 30.4 & 41.3\\
    w/ R.R. & 25.6 & 34.4 & 46.8 & 52.2 & 41.0 & 55.4 & 62.1 & 70.0 & 21.0 & 24.9 & 27.0 & 40.9\\
    \hline
    w/ Ours & \textbf{33.3} & \textbf{43.5} & \textbf{49.4} & \textbf{54.4} & \textbf{49.0} & \textbf{60.0} & \textbf{64.7} & \textbf{71.5} & \textbf{24.1} & \textbf{27.3} & \textbf{33.5} & \textbf{43.0}\\
    \rowcolor{gray!10}
    $\Delta$ & +11.0 & +8.7 & +5.9 & +2.8 & +14.1 & +8.3 & +7.3 & +5.5 & +5.2 & +6.1 & +7.8 & +7.6\\

    \hline
    \hline
    TCA-GCN~\cite{wangSkeletonbasedActionRecognition2022} & 29.8 & 39.6 & 46.1 & 52.7 & 40.9 & 52.9 & 59.8 & 67.3 & 20.9 & 24.7 & 27.1 & 33.8\\
    w/ Mixup & 33.0 & 42.0 & 46.9 & 52.5 & 45.8 & 57.6 & 62.5 & 68.8 & 23.6 & 27.2 & 29.3 & 42.1\\
    w/ R.R. & 27.9 & 41.7 & 47.4 & 53.4 & 46.0 & 57.4 & 63.5 & 70.7 & 24.8 & 29.6 & 32.0 & 37.9\\
    \hline
    w/ Ours & \textbf{35.4} & \textbf{42.8} & \textbf{49.5} & \textbf{54.8} & \textbf{51.3} & \textbf{61.7} & \textbf{67.3} & \textbf{73.1} & \textbf{26.1} & \textbf{30.0} & \textbf{35.4} & \textbf{43.7}\\
    \rowcolor{gray!10}
    $\Delta$ & +5.6 & +3.2 & +3.4 & +2.1 & +10.4 & +8.8 & +7.5 & +5.8 & +5.2 & +5.3 & +8.3 & +9.9\\
    \hline
    \hline
    BlockGCN~\cite{zhouBlockGCNRedefineTopology2024} & 25.5 & 33.8 & 38.2 & 47.9 & 34.8 & 48.1 & 53.7 & 61.0 & 18.3 & 24.3 & 28.2 & 39.5\\
    w/ Mixup & 28.3 & 34.8 & 41.6 & 47.7 & 38.5 & 53.4 & 58.2 & 66.5 & 20.4 & 28.7 & 34.2 & 44.2\\
    w/ R.R. & 27.5 & 31.4 & 40.2 & 46.5 & 37.3 & 51.1 & 56.0 & 64.0 & 19.8 & 29.1 & 34.2 & 41.2\\
    \hline
    w/ Ours & \textbf{30.4} & \textbf{37.2} & \textbf{43.8} & \textbf{49.5} & \textbf{45.2} & \textbf{56.8} & \textbf{63.1} & \textbf{68.7} & \textbf{25.0} & \textbf{34.0} & \textbf{39.4} & \textbf{47.4}\\
    \rowcolor{gray!10}
    $\Delta$ & +4.9 & +3.4 & +5.6 & +1.6 & +10.4 & +8.7 & +9.4 & +7.7 & +6.7 & +9.7 & +11.2 & +7.9\\
    \hline
    \hline
    CTR-GCN~\cite{chenChannelwiseTopologyRefinement2021} & 29.2 & 40.1 & 46.7 & 53.2 & 41.5 & 53.9 & 60.0 & 67.5 & 21.3 & 25.5 & 29.8 & 38.6\\
    w/ Mixup & 34.2 & 42.7 & 48.0 & 54.2 & 45.4 & 57.3 & 64.0 & 69.9 & 24.7 & 28.3 & 34.4 & 42.7\\
    w/ R.R. & 31.6 & 41.2 & 47.7 & 53.7 & 46.4 & 58.5 & 64.4 & 70.8 & 24.8 & 30.5 & 34.0 & 42.4\\
    \hline
    w/ Ours & \textbf{36.9} & \textbf{42.6} & \textbf{49.0} & \textbf{54.5} & \textbf{52.2} & \textbf{61.7} & \textbf{67.0} & \textbf{73.0} & \textbf{26.1} & \textbf{32.3} & \textbf{36.8} & \textbf{43.6}\\
    \rowcolor{gray!10}
    $\Delta$ & +7.7 & +2.5 & +2.3 & +1.3 & +10.7 & +7.8 & +7.0 & +5.5 & +4.8 & +6.8 & +7.0 & +5.0\\
  \hline
  \end{tabularx}
  \label{tab:comp_multi_backbone}
\end{table*}

\subsection{Ablation Studies}

\textbf{Ablation on Major Components.} Ablation studies are conducted under two low-data conditions: limited-scale setting with 10 samples per class and one-shot learning setting with 20 training classes to demonstrate the efficacy of each component under extreme conditions. The results are shown in Table~\ref{tab:abl_limit}. The sample pairs selection strategy along improves performance by encouraging the model to learn correlations. Subsequently, integrating the feature disentanglement module into the model results in a more substantial performance boost. By further introducing cross-sample aggregation, the model explicitly learns from the mixture of different samples, thereby increasing diversity and improving the performance. Each component contributes to the overall performance gains, demonstrating the effectiveness of our proposed method.

\textbf{Ablation on Hyper-parameters.} We analyze the balance parameter $w_x$ in the sample aggregation loss, with result shown in Table~\ref{tab:abl_hyper}. We find that $w_x=0.1$ yields better performance across all settings in limited-scale training. Nevertheless, the sample aggregation is still useful. As shown in Table~\ref{tab:abl_limit}, the ``+~Disent.'' column equivalent to setting this hyper-parameter to 0, reveals a notable performance deterioration to 47.2\%.

\textbf{Effectiveness of Feature Disentanglement.} To evaluate the contribution of the disentangled features to the final prediction, and demonstrate that the action feature and the performer feature are decoupled, we manually mask the action feature or performer feature by setting it to zero before aggregation. Results shown in Table~\ref{tab:abl_decouple} show a significant performance decline without the action feature. The absence of the performer feature impairs the performance but still surpasses the baseline. This indicates that the action-related feature predominantly drives the classification, yet still requires the performer feature for optimal performance. The performer-related feature appears to enhance action diversity, thus contributing to more accurate inference.

\textbf{Training Classes in One-shot Setting.} One important question in one-shot action recognition is how many training classes are sufficient for learning effective action representations. Following previous studies on NTU RGB+D 120 dataset, we train our model on different numbers of training classes. The result shown in Table~\ref{tab:os_cls} illustrates the competitive performance of our method, especially when training data is scarce.

\textbf{Ablation on Aggregation Strategy.}
To examine the impact of different designs on feature aggregation. We conduct ablation studies on different feature aggregation strategies in the limited-scale setting, presented in Table~\ref{tab:abl_feat_aggr} with the following setups:

(1) Concatenation: where disentangled features are first processed through an average pooling layer and then concatenated.

(2) Matrix Multiplication: which involves element-wise multiplication along the channel dimension, followed by a 1$\times$1 convolution and global average pooling across the temporal and spatial dimensions.

(3) Cross Attention: where the temporal feature serves as the query, and the spatial features act as key and value.

While other strategies slightly outperform concatenation in some settings on the NTU RGB+D dataset, the concatenation method consistently delivers more stable performance across different datasets. Furthermore, concatenation introduces no extra parameters and avoids heavy computation overhead. These findings suggest that cross-sample feature aggregation may enhance spatial and temporal modeling within GCN encoders, thus improving feature representation. Additionally, in limited-scale settings where the number of samples is small, aggregation methods like the attention mechanism may focus on noise, which could impair the robustness of the model on full-scale testing data. Since the main goal of our method is to learn better representations from scarce data, we adopt a simpler aggregation strategy, following Occam’s Razor, to prevent over-complicating the model.

\begin{table}
  \centering
  \renewcommand{\arraystretch}{1.2}
  \caption{Accuracy (\%) comparison on \textbf{one-shot} skeleton action recognition on NTU RGB+D and NTU RGB+D 120 datasets.}
  \begin{tabularx}{\columnwidth}{l*{3}{>{\centering\arraybackslash}X}}
    \hline
    Method & Venue & NTU & NTU-120\\
    \hline
    \textbf{Meta Learning} & & & \\
    ProtoNet~\cite{snellPrototypicalNetworksFewshot2017} & NIPS 2017 & 75.0 & 66.0 \\
    FEAT~\cite{yeFewShotLearningEmbedding2020} & CVPR 2020 & 80.5 & 65.1\\
    \hline
    \textbf{Base GCN Models} & & & \\
    ST-GCN~\cite{yanSpatialTemporalGraph2018} & AAAI 2018 & 71.7 & 63.3\\
    CTR-GCN~\cite{chenChannelwiseTopologyRefinement2021} & CVPR 2021 & 76.6 & 64.6\\
    TCA-GCN~\cite{wangSkeletonbasedActionRecognition2022} & Arxiv 2022 & 71.0 & 66.3\\
    \hline
    \textbf{One-shot Methods} & & &\\
    APSR~\cite{liuNTURGB1202020} & TPAMI 2020 & - & 45.3\\
    TCN~\cite{sabaterOneShotActionRecognition2021} & CVPR 2021 & - & 46.5\\
    SL-DML~\cite{memmesheimerSLDMLSignalLevel2021} & ICPR 2021 & - & 50.9\\
    Skeleton-DML~\cite{memmesheimerSkeletonDMLDeepMetric2022} & WACV 2022 & - & 54.2\\
    uDTW~\cite{wangUncertaintyDTWTimeSeries2022} & ECCV 2022 & 72.4 & 49.0\\
    JEANIE~\cite{wangTemporalViewpointTransportationPlan2022} & ACCV 2022 & 80.0 & 57.0\\
    ALCA-GCN~\cite{zhuAdaptiveLocalComponentawareGraph2023} & WACV 2023 & - & 57.6\\
    Koopman~\cite{wangNeuralKoopmanPooling2023} & CVPR 2023 & - & 68.1\\
    SMAM~\cite{liSMAMSelfMutual2023} & TIP 2023 & 73.6 & 56.4\\
    Trans4SOAR~\cite{peng2023delving} & TMM 2023 & 74.2 & 57.1\\
    STA-MLN~\cite{liSpatialTemporalAdaptiveMetric2024} & SPL 2024 & - & 59.9\\
    CrossGLG~\cite{yanCrossGLGLLMGuides2024a} & ECCV 2024 & 75.6 & 62.6\\
    M\&C-scale~\cite{yangOneShotActionRecognition2024} & TPAMI 2024 & 82.7 & 68.7\\
    \hline
    SkeletonX~(Ours) & - &  \textbf{83.2} & \textbf{69.1}\\
  \hline
  \end{tabularx}
  \label{tab:os_full_ntu}
\end{table}

\begin{table}
  \centering
  \renewcommand{\arraystretch}{1.2}
  \caption{Accuracy (\%) comparison on \textbf{one-shot} skeleton action recognition methods on PKU-MMD (II) dataset.}
  \begin{tabularx}{\columnwidth}{l*{2}{>{\centering\arraybackslash}X}}
    \hline
    Method & Venue & PKU-MMD (II)\\
    \hline
    \textbf{Meta Learning} & & \\
    ProtoNet~\cite{snellPrototypicalNetworksFewshot2017} & NIPS 2017 & 37.0 \\
    FEAT~\cite{yeFewShotLearningEmbedding2020} & CVPR 2020 & 32.3 \\
    \hline
    \textbf{Base GCN models} & & \\
    ST-GCN~\cite{yanSpatialTemporalGraph2018} & AAAI 2018 & 31.6\\
    CTR-GCN~\cite{chenChannelwiseTopologyRefinement2021} & CVPR 2022 & 32.5\\
    TCA-GCN~\cite{wangSkeletonbasedActionRecognition2022} & Arxiv 2023 & 28.8\\
    \hline
    SkeletonX~(Ours) & - & \textbf{38.3}\\
  \hline
  \end{tabularx}
  \label{tab:os_full_pku}
\end{table}

\begin{table*}[tb]
  \centering
  \small
  \renewcommand{\arraystretch}{1.2}
  \caption{Top-1 accuracy (\%) when combining our proposed method to transformer-based method SkateFormer~\cite{do2024skateformer} in \textbf{limited-scale setting}. The best top-1 accuracy is in \textbf{bold}.}
  \begin{tabularx}{\textwidth}{l|>{\centering\arraybackslash}X>{\centering\arraybackslash}X>{\centering\arraybackslash}X>{\centering\arraybackslash}X|>{\centering\arraybackslash}X>{\centering\arraybackslash}X>{\centering\arraybackslash}X>{\centering\arraybackslash}X|>{\centering\arraybackslash}X>{\centering\arraybackslash}X>{\centering\arraybackslash}X>{\centering\arraybackslash}X}
    \hline
    Methods & \multicolumn{4}{c|}{PKU-MMD (II)} & \multicolumn{4}{c|}{NTU} & \multicolumn{4}{c}{NTU-120}\\
     & 10 & 20 & 30 & 50 & 10 & 20 & 30 & 50 & 10 & 20 & 30 & 50\\
    \hline
    \hline
    SkateFormer~\cite{do2024skateformer} & 26.1 & 33.5 & 40.0 & 47.9 & 39.8 & 51.9 & 60.6 & 70.2 & 14.0 & \textbf{21.4} & \textbf{23.6} & 27.2\\
    w/ Ours & \textbf{28.2} & \textbf{38.2} & \textbf{46.8} & \textbf{55.7} & \textbf{44.4} & \textbf{58.9} & \textbf{66.2} & \textbf{72.8} & \textbf{16.4} & 20.9 & 22.2 & \textbf{29.7}\\
    \hline
    \rowcolor{gray!10}
    $\Delta$ & +2.1 & +4.7 & +6.8 & +7.8 & +4.6 & +7.0 & +5.6 & +2.6 & +2.4 & -0.5 & -1.4 & +2.5\\
  \hline
  \end{tabularx}
  \label{tab:more_backbones_trans}
\end{table*}

\subsection{Model Efficiency Comparison}
To illustrate the efficiency of our design, we compare it with two competitive methods. Since there is no official implementation code available for these methods, we obtain the code for their backbone models and calculate the number of parameters and FLOPs by running a single sample in inference mode. Then we calculate the additional parameters and FLOPs introduced by each method according to their papers and combine them with the baseline ones to get the final results. Due to the lightweight and universal module design of our method, the number of parameters only increases by approximately 0.07M. As presented in Table~\ref{tab:os_flops}, our lightweight design achieves competitive results with state-of-the-art methods~\cite{yangOneShotActionRecognition2024} with 1/10 parameters and 1/40 FLOPs.

\begin{table*}[tb]
  \centering
  \small
  \renewcommand{\arraystretch}{1.2}
  \caption{Extending our proposed method to \textbf{bone modality} in \textbf{limited-scale setting}. The best top-1 accuracy (\%) is in \textbf{bold}.}
  \begin{tabularx}{\textwidth}{l|>{\centering\arraybackslash}X>{\centering\arraybackslash}X>{\centering\arraybackslash}X>{\centering\arraybackslash}X|>{\centering\arraybackslash}X>{\centering\arraybackslash}X>{\centering\arraybackslash}X>{\centering\arraybackslash}X|>{\centering\arraybackslash}X>{\centering\arraybackslash}X>{\centering\arraybackslash}X>{\centering\arraybackslash}X}
    \hline
    Methods & \multicolumn{4}{c|}{PKU-MMD (II)} & \multicolumn{4}{c|}{NTU} & \multicolumn{4}{c}{NTU-120}\\
     & 10 & 20 & 30 & 50 & 10 & 20 & 30 & 50 & 10 & 20 & 30 & 50\\
    \hline
    \hline
    ST-GCN & 23.8 & 29.4 & 35.5 & 46.6 & 28.7 & 41.1 & 48.1 & 58.7 & 12.7 & 18.9 & 23.1 & 34.1\\
    w/ R.R. & 24.2 & 34.5 & 40.1 & 49.8 & 38.4 & 48.3 & 55.8 & 65.8 & 14.9 & 21.0 & 27.2 & 37.4\\
    \hline
    w/ Ours & \textbf{26.7} & \textbf{38.8} & \textbf{41.5} & \textbf{49.7} & \textbf{41.0} & \textbf{52.8} & \textbf{62.2} & \textbf{69.5} & \textbf{20.3} & \textbf{25.4} & \textbf{32.9} & \textbf{42.9}\\
    \rowcolor{gray!10}
    $\Delta$ & +2.9 & +9.4 & +6.0 & +3.1 & +12.3 & +11.7 & +14.1 & +10.8 & +7.6 & +6.5 & +9.8 & +8.8\\
    \hline
    \hline
    TCA-GCN & 25.5 & 32.5 & 41.4 & 49.3 & 32.2 & 45.5 & 52.9 & 63.4 & 13.1 & 19.4 & 24.6 & 38.0\\
    w/ R.R. & 24.8 & 35.5 & 42.2 & 51.9 & 40.8 & 54.4 & 60.1 & 69.0 & 16.0 & 20.7 & 27.9 & 41.8\\
    \hline
    w/ Ours & \textbf{31.5} & \textbf{40.7} & \textbf{47.2} & \textbf{54.7} & \textbf{46.5} & \textbf{58.6} & \textbf{65.3} & \textbf{72.3} & \textbf{20.5} & \textbf{28.3} & \textbf{37.0} & \textbf{45.1}\\
    $\Delta$ & +6.0 & +8.2 & +5.8 & +5.4 & +14.3 & +13.1 & +12.4 & +8.9 & +7.4 & +9.1 & +12.4 & +7.1\\
    \hline
    \hline
    CTR-GCN & 26.0 & 33.7 & 41.4 & 48.6 & 30.9 & 46.4 & 54.4 & 65.1 & 15.8 & 21.9 & 25.7 & 37.8\\
    w/ R.R. & 25.6 & 37.4 & 44.5 & 51.9 & 40.4 & 54.4 & 61.3 & 69.6 & 17.6 & 25.0 & 31.6 & 40.4\\
    \hline
    w/ Ours & \textbf{29.4} & \textbf{39.8} & \textbf{46.7} & \textbf{54.1} & \textbf{45.9} & \textbf{58.0} & \textbf{65.5} & \textbf{72.0} & \textbf{22.9} & \textbf{30.6} & \textbf{36.9} & \textbf{45.2}\\
    \rowcolor{gray!10}
    $\Delta$ & +3.4 & +6.1 & +5.3 & +5.5 & +15.0 & +11.6 & +11.1 & +6.9 & +7.1 & +8.7 & +11.2 & +7.4\\
  \hline
  \end{tabularx}
  \label{tab:modality_bone}
\end{table*}

\begin{figure}[tb]
  \centering
  \includegraphics[width=\linewidth]{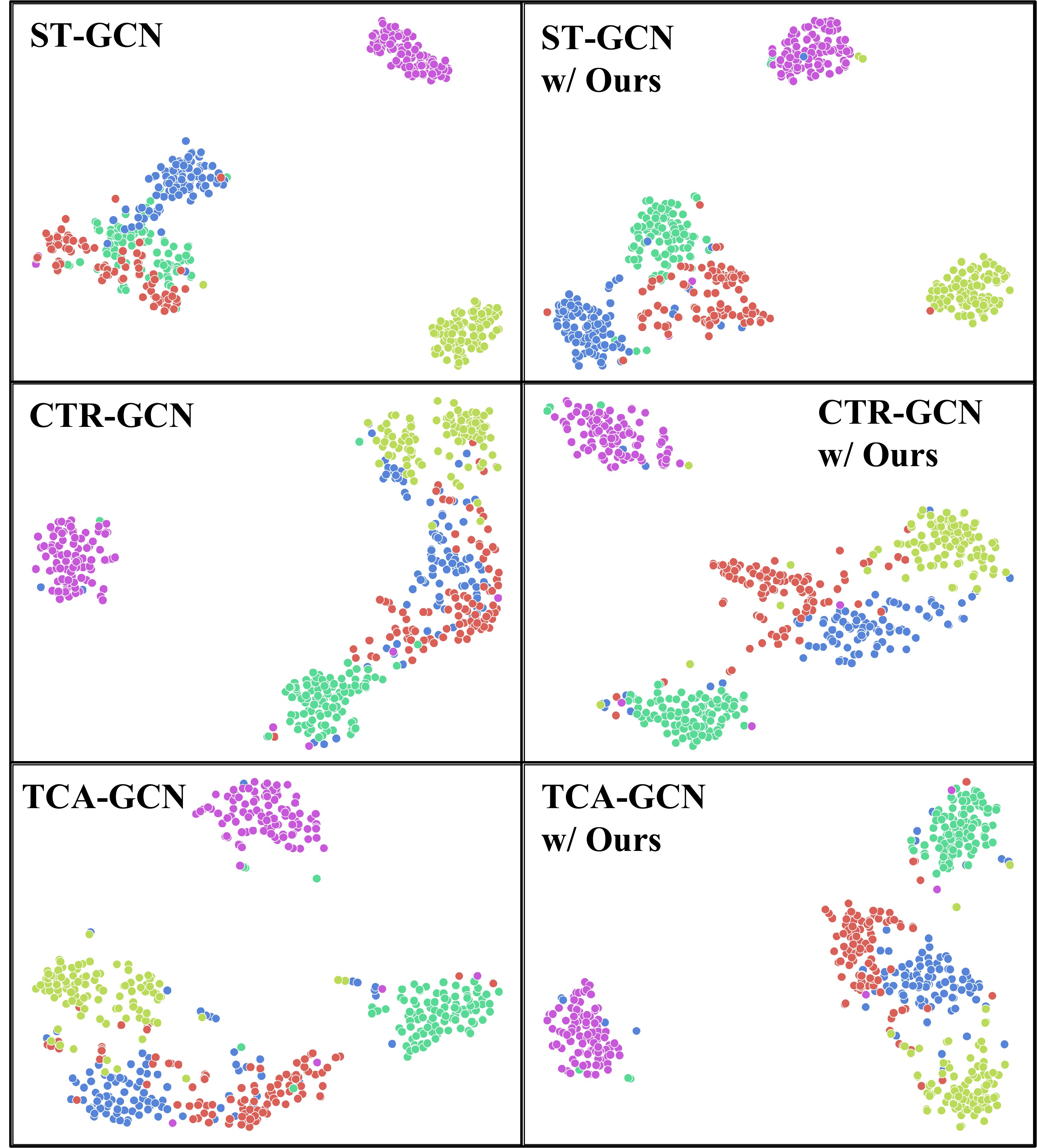}
  \caption{Visualization of feature space using t-SNE for samples from NTU RGB+D dataset. Each row represents a different GCN backbone. The first column presents the results from the original backbones, while the second column displays the results from our proposed method.}
\label{fig:tsne_viz}
\end{figure}

\subsection{Visualization on Action Feature}
To gain insight into the feature space distribution, we randomly select several categories from the NTU RGB+D dataset and visualize their distributions using t-SNE, as shown in Fig.~\ref{fig:tsne_viz}. Specifically, we extract the output feature before the FC layer for the original backbones, and the output action-related feature from the feature disentanglement module for our proposed method. Each column represents one GCN backbone. The upper row is from the original backbones, and the bottom row is from our proposed method. It can be observed that our method achieves a more discriminative feature representation, characterized by clearer boundaries and more compact clusters.

\subsection{Combine with Base Models}
To demonstrate the universality of our approach across different GCN architectures, we implement our method on various GCN models. For fair comparison, we re-implement the baseline methods and test them on multiple datasets under the limited-scale setting. The results are presented in Table~\ref{tab:comp_multi_backbone}. Compared to other augmentation methods in skeleton-based action recognition, our proposed method improves the performance the most across various backbones. The maximum improvement is 14.1\% in 10 samples per class setting, highlighting its effectiveness, particularly when training data are scarce.

When applied to transformer-based models, our method also shows performance gains with some settings, as illustrated in Table~\ref{tab:more_backbones_trans}. SkateFormer includes multiple data augmentation methods, we keep all of them as the baseline and implement our method. We observe that performance improvements are less significant compared to those achieved with GCN-based backbones. This discrepancy could be attributed to the inherent differences between the global attention mechanism of the Transformer and the topology-based approach of GCNs. The attention mechanism in Transformers may not perform as effectively in settings with fewer data, as it may focus on less relevant parts of the data.

\subsection{Applying to Other Modality}

To validate the effectiveness of our method on other modalities, we have extended our experiments to the bone modality, which is also widely used in skeleton-based action recognition~\cite{liuNTURGB1202020, chenChannelwiseTopologyRefinement2021, zhouLearningDiscriminativeRepresentations2023a}. The results, as shown in Table~\ref{tab:modality_bone}, demonstrate that our method performs well in this modality, with a significant performance increase of up to 15.0\% on the NTU RGB+D dataset with 10 samples per category.

Our approach consistently outperforms the baseline methods across different datasets, including PKU-MMD (II), NTU, and NTU-120. This highlights the versatility and robustness of our method in adapting to different input modalities and the limited-scale setting.

\begin{figure}[t]
\centering
\includegraphics[width=\linewidth]{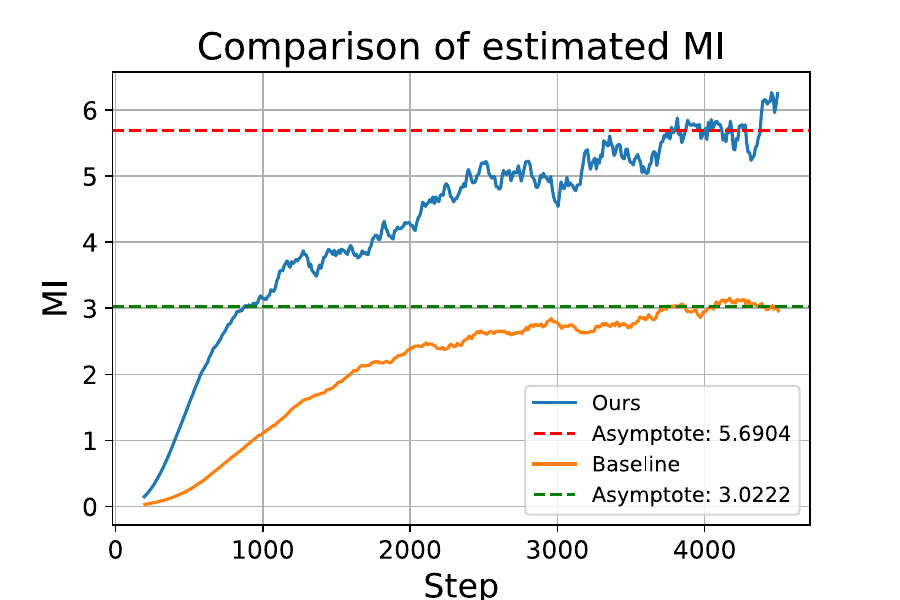}
\caption{Comparison on estimated Mutual Information (MI). The estimated MI converges to a stable value as the estimator is trained. Our method improves the mutual information compared to the baseline.}
\label{fig:mi_comparison}
\end{figure}

\subsection{Comparison With State-of-the-Art Methods}

To validate the effectiveness of our method with increasing pretraining data, we conduct a comprehensive comparison with state-of-the-art one-shot methods. We also re-implement several meta-learning methods compatible with GCNs for comparison. Results of NTU RGB+D and NTU RGB+D 120 are shown in Table~\ref{tab:os_full_ntu} and PKU-MMD (II) in Table~\ref{tab:os_full_pku}, demonstrating that our method achieves state-of-the-art performance on multiple datasets. Our method achieves competitive results to advanced matching-based approaches~\cite{wangNeuralKoopmanPooling2023, yangOneShotActionRecognition2024} while maintaining a concise and lightweight design.

\subsection{Mutual Information Estimation}
\label{sec:MI_estimation}
To validate the hypothesis that cross-sample feature aggregation improves Mutual Information (MI), we compare two MI values: the baseline MI, $I(F;Y)$, calculated using the original feature $F=f_\theta(X)$, and the MI, $I(V;Y)$, computed using the aggregated feature $V=A(F,F')$ in our proposed method. Since it is difficult to directly compute MI between variables of different dimensions, we estimate MI through MINE method~\cite{belghazi2021mine}.

Specifically, we use ST-GCN~\cite{yanSpatialTemporalGraph2018} as the baseline model and train the MI estimator for 4500 steps (500 epochs) on 10 samples per category from the NTU RGB+D dataset with a batch size of 64. As the MI estimator is trained, it gradually converges to a stable value. The relative size of this value allows us to compare the differences in mutual information between the baseline and our method. The results, shown in Figure~\ref{fig:mi_comparison}, indicate that our method improves the MI from 3.02 to 5.69, outperforming the baseline. This verifies the assumption in Section~\ref{sec:theory}

\section{Conclusion}
In this paper, we investigate two low-data scenarios for skeleton-based action recognition and propose a novel cross-sample aggregation method for action recognition under limited training data. First, it selects two types of sample pairs for each sample within a batch, then disentangles the sample and its pair to performer-related and action-related features. The decoupled features of the paired samples are crossed over and subsequently reassembled through an aggregation module. We also provide an analysis based on IB theory for a better understanding of the underlying mechanism of our proposed method. Our method can be easily integrated with multiple GCN backbones, demonstrating effectiveness and compatibility, and achieves competitive performance with fewer parameters and FLOPs.

\textbf{Discussion:} 1) While our method demonstrates strong performance on GCN-based models, the performance gains on transformer architectures remain unstable. Currently, our method is tailored for GCN models. We plan to further explore the two key attributes in transformer-based architectures, focusing on optimizing the aggregation process to improve stability and performance. 2) We have been able to provide formal proofs for optimizing one term in the IB theory, however, the second term remains unproven due to the complex nature of deep learning models and the limitations of multi-variable mutual information estimation techniques. The black-box nature of deep learning models further complicates the derivation of a comprehensive theoretical justification. We believe that with a more systematic and formalized proof, coupled with corresponding experiments, the cross-sample aggregation method can be extended beyond the current application to a broader range of fields.

In conclusion, we hope that our work contributes to the advancement of data-efficient and computationally efficient skeleton-based action recognition, and inspires applications in other domains.


\ifCLASSOPTIONcaptionsoff
  \newpage
\fi

\bibliographystyle{IEEEtran}
\bibliography{main}

\begin{thebibliography}{10}
\providecommand{\url}[1]{#1}
\csname url@samestyle\endcsname
\providecommand{\newblock}{\relax}
\providecommand{\bibinfo}[2]{#2}
\providecommand{\BIBentrySTDinterwordspacing}{\spaceskip=0pt\relax}
\providecommand{\BIBentryALTinterwordstretchfactor}{4}
\providecommand{\BIBentryALTinterwordspacing}{\spaceskip=\fontdimen2\font plus
\BIBentryALTinterwordstretchfactor\fontdimen3\font minus \fontdimen4\font\relax}
\providecommand{\BIBforeignlanguage}[2]{{%
\expandafter\ifx\csname l@#1\endcsname\relax
\typeout{** WARNING: IEEEtran.bst: No hyphenation pattern has been}%
\typeout{** loaded for the language `#1'. Using the pattern for}%
\typeout{** the default language instead.}%
\else
\language=\csname l@#1\endcsname
\fi
#2}}
\providecommand{\BIBdecl}{\relax}
\BIBdecl

\bibitem{pavlakosCoarseToFineVolumetricPrediction2017}
G.~Pavlakos, X.~Zhou, K.~G. Derpanis, and K.~Daniilidis, ``Coarse-{{To-Fine Volumetric Prediction}} for {{Single-Image 3D Human Pose}},'' in \emph{Proceedings of the {{IEEE Conference}} on {{Computer Vision}} and {{Pattern Recognition}}}, 2017, pp. 7025--7034.

\bibitem{caiExploitingSpatialTemporalRelationships2019}
Y.~Cai, L.~Ge, J.~Liu, J.~Cai, T.-J. Cham, J.~Yuan, and N.~M. Thalmann, ``Exploiting {{Spatial-Temporal Relationships}} for {{3D Pose Estimation}} via {{Graph Convolutional Networks}},'' in \emph{Proceedings of the {{IEEE}}/{{CVF International Conference}} on {{Computer Vision}}}, 2019, pp. 2272--2281.

\bibitem{zhaoSemanticGraphConvolutional2019}
L.~Zhao, X.~Peng, Y.~Tian, M.~Kapadia, and D.~N. Metaxas, ``Semantic {{Graph Convolutional Networks}} for {{3D Human Pose Regression}},'' in \emph{Proceedings of the {{IEEE}}/{{CVF Conference}} on {{Computer Vision}} and {{Pattern Recognition}}}, 2019, pp. 3425--3435.

\bibitem{zhao2024crosscamera}
Y.~Zhao, G.~Li, and E.~Y. Lam, ``Cross-{{Camera Human Motion Transfer}} by {{Time Series Analysis}},'' in \emph{{{ICASSP}} 2024 - 2024 {{IEEE International Conference}} on {{Acoustics}}, {{Speech}} and {{Signal Processing}} ({{ICASSP}})}, 2024-04, pp. 4985--4989.

\bibitem{shahroudyNTURGBLarge2016}
A.~Shahroudy, J.~Liu, T.-T. Ng, and G.~Wang, ``{{NTU RGB}}+{{D}}: {{A Large Scale Dataset}} for {{3D Human Activity Analysis}},'' in \emph{Proceedings of the {{IEEE Conference}} on {{Computer Vision}} and {{Pattern Recognition}}}, 2016, pp. 1010--1019.

\bibitem{liuNTURGB1202020}
J.~Liu, A.~Shahroudy, M.~Perez, G.~Wang, L.-Y. Duan, and A.~C. Kot, ``{{NTU RGB}}+{{D}} 120: {{A Large-Scale Benchmark}} for {{3D Human Activity Understanding}},'' \emph{IEEE Transactions on Pattern Analysis and Machine Intelligence}, vol.~42, no.~10, pp. 2684--2701, 2020.

\bibitem{liuBenchmarkDatasetComparison2020}
J.~Liu, S.~Song, C.~Liu, Y.~Li, and Y.~Hu, ``A {{Benchmark Dataset}} and {{Comparison Study}} for {{Multi-modal Human Action Analytics}},'' \emph{ACM Trans. Multimedia Comput. Commun. Appl.}, vol.~16, no.~2, pp. 41:1--41:24, 2020.

\bibitem{yanSpatialTemporalGraph2018}
S.~Yan, Y.~Xiong, and D.~Lin, ``Spatial {{Temporal Graph Convolutional Networks}} for {{Skeleton-Based Action Recognition}},'' \emph{AAAI}, vol.~32, no.~1, 2018.

\bibitem{shiTwoStreamAdaptiveGraph2019}
L.~Shi, Y.~Zhang, J.~Cheng, and H.~Lu, ``Two-{{Stream Adaptive Graph Convolutional Networks}} for {{Skeleton-Based Action Recognition}},'' in \emph{Proceedings of the {{IEEE}}/{{CVF Conference}} on {{Computer Vision}} and {{Pattern Recognition}}}, 2019, pp. 12\,026--12\,035.

\bibitem{liuDisentanglingUnifyingGraph2020}
Z.~Liu, H.~Zhang, Z.~Chen, Z.~Wang, and W.~Ouyang, ``Disentangling and {{Unifying Graph Convolutions}} for {{Skeleton-Based Action Recognition}},'' in \emph{Proceedings of the {{IEEE}}/{{CVF Conference}} on {{Computer Vision}} and {{Pattern Recognition}}}, 2020, pp. 143--152.

\bibitem{chengDecouplingGCNDropGraph2020}
K.~Cheng, Y.~Zhang, C.~Cao, L.~Shi, J.~Cheng, and H.~Lu, ``Decoupling {{GCN}} with {{DropGraph Module}} for {{Skeleton-Based Action Recognition}},'' in \emph{Computer {{Vision}} -- {{ECCV}} 2020}, ser. Lecture {{Notes}} in {{Computer Science}}, A.~Vedaldi, H.~Bischof, T.~Brox, and J.-M. Frahm, Eds.\hskip 1em plus 0.5em minus 0.4em\relax Cham: Springer International Publishing, 2020, pp. 536--553.

\bibitem{korbanDDGCNDynamicDirected2020}
M.~Korban and X.~Li, ``{{DDGCN}}: {{A Dynamic Directed Graph Convolutional Network}} for {{Action Recognition}},'' in \emph{Computer {{Vision}} -- {{ECCV}} 2020}, ser. Lecture {{Notes}} in {{Computer Science}}, A.~Vedaldi, H.~Bischof, T.~Brox, and J.-M. Frahm, Eds.\hskip 1em plus 0.5em minus 0.4em\relax Cham: Springer International Publishing, 2020, pp. 761--776.

\bibitem{chenChannelwiseTopologyRefinement2021}
Y.~Chen, Z.~Zhang, C.~Yuan, B.~Li, Y.~Deng, and W.~Hu, ``Channel-wise {{Topology Refinement Graph Convolution}} for {{Skeleton-Based Action Recognition}},'' in \emph{2021 {{IEEE}}/{{CVF International Conference}} on {{Computer Vision}} ({{ICCV}})}.\hskip 1em plus 0.5em minus 0.4em\relax Montreal, QC, Canada: IEEE, 2021, pp. 13\,339--13\,348.

\bibitem{wangSkeletonbasedActionRecognition2022}
S.~Wang, Y.~Zhang, M.~Zhao, H.~Qi, K.~Wang, F.~Wei, and Y.~Jiang, ``Skeleton-based {{Action Recognition}} via {{Temporal-Channel Aggregation}},'' https://arxiv.org/abs/2205.15936v2, 2022.

\bibitem{leeHierarchicallyDecomposedGraph2023}
J.~Lee, M.~Lee, D.~Lee, and S.~Lee, ``Hierarchically {{Decomposed Graph Convolutional Networks}} for {{Skeleton-Based Action Recognition}},'' in \emph{Proceedings of the {{IEEE}}/{{CVF International Conference}} on {{Computer Vision}}}, 2023, pp. 10\,444--10\,453.

\bibitem{zhouBlockGCNRedefineTopology2024}
Y.~Zhou, X.~Yan, Z.-Q. Cheng, Y.~Yan, Q.~Dai, and X.-S. Hua, ``{{BlockGCN}}: {{Redefine Topology Awareness}} for {{Skeleton-Based Action Recognition}},'' in \emph{Proceedings of the {{IEEE}}/{{CVF Conference}} on {{Computer Vision}} and {{Pattern Recognition}}}, 2024, pp. 2049--2058.

\bibitem{zhangCrossScaleSpatiotemporalRefinement2024}
Y.~Zhang, Z.~Sun, M.~Dai, J.~Feng, and K.~Jia, ``Cross-{{Scale Spatiotemporal Refinement Learning}} for {{Skeleton-Based Action Recognition}},'' \emph{IEEE Signal Process. Lett.}, vol.~31, pp. 441--445, 2024.

\bibitem{wang2024survey}
\BIBentryALTinterwordspacing
K.~Wang, J.~Zhu, M.~Ren, Z.~Liu, S.~Li, Z.~Zhang, C.~Zhang, X.~Wu, Q.~Zhan, Q.~Liu, and Y.~Wang, ``A survey on data synthesis and augmentation for large language models.'' [Online]. Available: \url{http://arxiv.org/abs/2410.12896}
\BIBentrySTDinterwordspacing

\bibitem{memmesheimerSkeletonDMLDeepMetric2022}
R.~Memmesheimer, S.~Haring, N.~Theisen, and D.~Paulus, ``Skeleton-{{DML}}: {{Deep Metric Learning}} for {{Skeleton-Based One-Shot Action Recognition}},'' in \emph{2022 {{IEEE}}/{{CVF Winter Conference}} on {{Applications}} of {{Computer Vision}} ({{WACV}})}.\hskip 1em plus 0.5em minus 0.4em\relax Waikoloa, HI, USA: IEEE, 2022, pp. 837--845.

\bibitem{wangTemporalViewpointTransportationPlan2022}
L.~Wang and P.~Koniusz, ``Temporal-{{Viewpoint Transportation Plan}} for {{Skeletal Few-shot Action Recognition}},'' in \emph{Proceedings of the {{Asian Conference}} on {{Computer Vision}}}, 2022, pp. 4176--4193.

\bibitem{yangOneShotActionRecognition2024}
S.~Yang, J.~Liu, S.~Lu, E.~M. Hwa, and A.~C. Kot, ``One-{{Shot Action Recognition}} via {{Multi-Scale Spatial-Temporal Skeleton Matching}},'' \emph{IEEE Transactions on Pattern Analysis and Machine Intelligence}, pp. 1--8, 2024.

\bibitem{caoRealtimeMultiPerson2D2017}
Z.~Cao, T.~Simon, S.-E. Wei, and Y.~Sheikh, ``Realtime {{Multi-Person 2D Pose Estimation Using Part Affinity Fields}},'' in \emph{Proceedings of the {{IEEE Conference}} on {{Computer Vision}} and {{Pattern Recognition}}}, 2017, pp. 7291--7299.

\bibitem{kayKineticsHumanAction2017}
W.~Kay, J.~Carreira, K.~Simonyan, B.~Zhang, C.~Hillier, S.~Vijayanarasimhan, F.~Viola, T.~Green, T.~Back, P.~Natsev, M.~Suleyman, and A.~Zisserman, ``The {{Kinetics Human Action Video Dataset}},'' 2017.

\bibitem{sabaterOneShotActionRecognition2021}
A.~Sabater, L.~Santos, J.~{Santos-Victor}, A.~Bernardino, L.~Montesano, and A.~C. Murillo, ``One-{{Shot Action Recognition}} in {{Challenging Therapy Scenarios}},'' in \emph{Proceedings of the {{IEEE}}/{{CVF Conference}} on {{Computer Vision}} and {{Pattern Recognition}}}, 2021, pp. 2777--2785.

\bibitem{deyzelOneshotSkeletonbasedAction2023}
M.~Deyzel and R.~P. Theart, ``One-shot skeleton-based action recognition on strength and conditioning exercises,'' in \emph{2023 {{IEEE}}/{{CVF Conference}} on {{Computer Vision}} and {{Pattern Recognition Workshops}} ({{CVPRW}})}.\hskip 1em plus 0.5em minus 0.4em\relax Vancouver, BC, Canada: IEEE, 2023, pp. 5169--5178.

\bibitem{zhangMixupEmpiricalRisk2018}
H.~Zhang, M.~Cisse, Y.~N. Dauphin, and D.~{Lopez-Paz}, ``Mixup: {{Beyond Empirical Risk Minimization}},'' 2018.

\bibitem{mengSampleFusionNetwork2019}
F.~Meng, H.~Liu, Y.~Liang, J.~Tu, and M.~Liu, ``Sample {{Fusion Network}}: {{An End-to-End Data Augmentation Network}} for {{Skeleton-Based Human Action Recognition}},'' \emph{IEEE Trans. on Image Process.}, vol.~28, no.~11, pp. 5281--5295, 2019.

\bibitem{zhang2025skeletonmix}
Z.~Zhang, H.~Zhou, Q.~Liu, and Y.~Wang, ``Skeletonmix: A {Mixup}-{Based} {Data} {Augmentation} {Framework} for {Skeleton}-{Based} {Action} {Recognition},'' in \emph{IEEE {International} {Conference} on {Acoustics}, {Speech}, and {Signal} {Processing}}, 2025.

\bibitem{tishby2000information}
\BIBentryALTinterwordspacing
N.~Tishby, F.~C. Pereira, and W.~Bialek. (2000-04-24) The information bottleneck method. [Online]. Available: \url{http://arxiv.org/abs/physics/0004057}
\BIBentrySTDinterwordspacing

\bibitem{alemi2017deep}
A.~A. Alemi, I.~Fischer, J.~V. Dillon, and K.~Murphy, ``Deep {Variational} {Information} {Bottleneck}.'' in \emph{International {Conference} on {Learning} {Representations}}, 2017.

\bibitem{duHierarchicalRecurrentNeural2015}
Y.~Du, W.~Wang, and L.~Wang, ``Hierarchical {{Recurrent Neural Network}} for {{Skeleton Based Action Recognition}},'' in \emph{Proceedings of the {{IEEE Conference}} on {{Computer Vision}} and {{Pattern Recognition}}}, 2015, pp. 1110--1118.

\bibitem{zhangGeometricFeaturesSkeletonBased2017}
S.~Zhang, X.~Liu, and J.~Xiao, ``On {{Geometric Features}} for {{Skeleton-Based Action Recognition Using Multilayer LSTM Networks}},'' in \emph{2017 {{IEEE Winter Conference}} on {{Applications}} of {{Computer Vision}} ({{WACV}})}, 2017, pp. 148--157.

\bibitem{zhangFusingGeometricFeatures2018}
S.~Zhang, Y.~Yang, J.~Xiao, X.~Liu, Y.~Yang, D.~Xie, and Y.~Zhuang, ``Fusing {{Geometric Features}} for {{Skeleton-Based Action Recognition Using Multilayer LSTM Networks}},'' \emph{IEEE Transactions on Multimedia}, vol.~20, no.~9, pp. 2330--2343, 2018.

\bibitem{liuSkeletonBasedHumanAction2018}
J.~Liu, G.~Wang, L.-Y. Duan, K.~Abdiyeva, and A.~C. Kot, ``Skeleton-{{Based Human Action Recognition With Global Context-Aware Attention LSTM Networks}},'' \emph{IEEE Transactions on Image Processing}, vol.~27, no.~4, pp. 1586--1599, 2018.

\bibitem{liActionalStructuralGraphConvolutional2019}
M.~Li, S.~Chen, X.~Chen, Y.~Zhang, Y.~Wang, and Q.~Tian, ``Actional-{{Structural Graph Convolutional Networks}} for {{Skeleton-Based Action Recognition}},'' in \emph{Proceedings of the {{IEEE}}/{{CVF Conference}} on {{Computer Vision}} and {{Pattern Recognition}}}, 2019, pp. 3595--3603.

\bibitem{tian2023skeletonbased}
H.~Tian, X.~Ma, X.~Li, and Y.~Li, ``Skeleton-{Based} {Action} {Recognition} {With} {Select}-{Assemble}-{Normalize} {Graph} {Convolutional} {Networks}.'' \emph{IEEE Transactions on Multimedia (TMM)}, vol.~25, pp. 8527--8538, 2023.

\bibitem{zhuAdaptiveLocalComponentawareGraph2023}
A.~Zhu, Q.~Ke, M.~Gong, and J.~Bailey, ``Adaptive {{Local-Component-aware Graph Convolutional Network}} for {{One-shot Skeleton-based Action Recognition}},'' in \emph{2023 {{IEEE}}/{{CVF Winter Conference}} on {{Applications}} of {{Computer Vision}} ({{WACV}})}.\hskip 1em plus 0.5em minus 0.4em\relax Waikoloa, HI, USA: IEEE, 2023, pp. 6027--6036.

\bibitem{peng2023delving}
K.~Peng, A.~Roitberg, K.~Yang, J.~Zhang, and R.~Stiefelhagen, ``Delving {Deep} {Into} {One}-{Shot} {Skeleton}-{Based} {Action} {Recognition} {With} {Diverse} {Occlusions}.'' \emph{IEEE Transactions on Multimedia (TMM)}, vol.~25, pp. 1489--1504, 2023.

\bibitem{maLearningSpatialPreservedSkeleton2022}
N.~Ma, H.~Zhang, X.~Li, S.~Zhou, Z.~Zhang, J.~Wen, H.~Li, J.~Gu, and J.~Bu, ``Learning {{Spatial-Preserved Skeleton Representations}} for~{{Few-Shot Action Recognition}},'' in \emph{Computer {{Vision}} -- {{ECCV}} 2022}, ser. Lecture {{Notes}} in {{Computer Science}}, S.~Avidan, G.~Brostow, M.~Ciss{\'e}, G.~M. Farinella, and T.~Hassner, Eds.\hskip 1em plus 0.5em minus 0.4em\relax Cham: Springer Nature Switzerland, 2022, pp. 174--191.

\bibitem{liuParallelAttentionInteraction2023}
X.~Liu, S.~Zhou, L.~Wang, and G.~Hua, ``Parallel {{Attention Interaction Network}} for {{Few-Shot Skeleton-Based Action Recognition}},'' in \emph{Proceedings of the {{IEEE}}/{{CVF International Conference}} on {{Computer Vision}}}, 2023, pp. 1379--1388.

\bibitem{duan2022dgstgcn}
H.~Duan, J.~Wang, K.~Chen, and D.~Lin, ``Dg-{STGCN}: Dynamic {Spatial}-{Temporal} {Modeling} for {Skeleton}-based {Action} {Recognition},'' \emph{arXiv}, vol. abs/2210.05895, 2022.

\bibitem{chi2022infogcn}
H.-G. Chi, M.~H. Ha, S.~Chi, S.~W. Lee, Q.~Huang, and K.~Ramani, ``{{InfoGCN}}: {{Representation Learning}} for {{Human Skeleton-based Action Recognition}},'' in \emph{2022 {{IEEE}}/{{CVF Conference}} on {{Computer Vision}} and {{Pattern Recognition}} ({{CVPR}})}.\hskip 1em plus 0.5em minus 0.4em\relax IEEE, 2022-06, pp. 20\,154--20\,164.

\bibitem{memmesheimerSLDMLSignalLevel2021}
R.~Memmesheimer, N.~Theisen, and D.~Paulus, ``{{SL-DML}}: {{Signal Level Deep Metric Learning}} for {{Multimodal One-Shot Action Recognition}},'' in \emph{2020 25th {{International Conference}} on {{Pattern Recognition}} ({{ICPR}})}, 2021, pp. 4573--4580.

\bibitem{wangUncertaintyDTWTimeSeries2022}
L.~Wang and P.~Koniusz, ``Uncertainty-{{DTW}} for~{{Time Series}} and~{{Sequences}},'' in \emph{Computer {{Vision}} -- {{ECCV}} 2022}, ser. Lecture {{Notes}} in {{Computer Science}}, S.~Avidan, G.~Brostow, M.~Ciss{\'e}, G.~M. Farinella, and T.~Hassner, Eds.\hskip 1em plus 0.5em minus 0.4em\relax Cham: Springer Nature Switzerland, 2022, pp. 176--195.

\bibitem{liSMAMSelfMutual2023}
Z.~Li, X.~Gong, R.~Song, P.~Duan, J.~Liu, and W.~Zhang, ``{{SMAM}}: {{Self}} and {{Mutual Adaptive Matching}} for {{Skeleton-Based Few-Shot Action Recognition}},'' \emph{IEEE Transactions on Image Processing}, vol.~32, pp. 392--402, 2023.

\bibitem{liSpatialTemporalAdaptiveMetric2024}
X.~Li, J.~Lu, X.~Chen, and X.~Zhang, ``Spatial-{{Temporal Adaptive Metric Learning Network}} for {{One-Shot Skeleton-Based Action Recognition}},'' \emph{IEEE Signal Processing Letters}, vol.~31, pp. 321--325, 2024.

\bibitem{yanCrossGLGLLMGuides2024a}
T.~Yan, W.~Zeng, Y.~Xiao, X.~Tong, B.~Tan, Z.~Fang, Z.~Cao, and J.~T. Zhou, ``{{CrossGLG}}: {{LLM Guides One-shot Skeleton-based 3D Action Recognition}} in a {{Cross-level Manner}},'' 2024.

\bibitem{do2024skateformer}
J.~Do and M.~Kim, ``Skateformer: Skeletal-{Temporal} {Transformer} for {Human} {Action} {Recognition}.'' in \emph{European {Conference} on {Computer} {Vision} ({ECCV})}, 2024, pp. 401--420.

\bibitem{snellPrototypicalNetworksFewshot2017}
J.~Snell, K.~Swersky, and R.~Zemel, ``Prototypical {{Networks}} for {{Few-shot Learning}},'' in \emph{Advances in {{Neural Information Processing Systems}}}, vol.~30.\hskip 1em plus 0.5em minus 0.4em\relax Curran Associates, Inc., 2017.

\bibitem{yeFewShotLearningEmbedding2020}
H.-J. Ye, H.~Hu, D.-C. Zhan, and F.~Sha, ``Few-{{Shot Learning}} via {{Embedding Adaptation With Set-to-Set Functions}},'' in \emph{Proceedings of the {{IEEE}}/{{CVF Conference}} on {{Computer Vision}} and {{Pattern Recognition}}}, 2020, pp. 8808--8817.

\bibitem{longSTEPCATFormerSpatialTemporal2023}
N.~H.~B. Long, ``{{STEP CATFormer}}: {{Spatial-Temporal Effective Body-Part Cross Attention Transformer}} for {{Skeleton-based Action Recognition}},'' 2023.

\bibitem{zhouLearningDiscriminativeRepresentations2023a}
H.~Zhou, Q.~Liu, and Y.~Wang, ``Learning {{Discriminative Representations}} for {{Skeleton Based Action Recognition}},'' in \emph{Proceedings of the {{IEEE}}/{{CVF Conference}} on {{Computer Vision}} and {{Pattern Recognition}}}, 2023, pp. 10\,608--10\,617.

\bibitem{wangNeuralKoopmanPooling2023}
X.~Wang, X.~Xu, and Y.~Mu, ``Neural {{Koopman Pooling}}: {{Control-Inspired Temporal Dynamics Encoding}} for {{Skeleton-Based Action Recognition}},'' in \emph{2023 {{IEEE}}/{{CVF Conference}} on {{Computer Vision}} and {{Pattern Recognition}} ({{CVPR}})}.\hskip 1em plus 0.5em minus 0.4em\relax Vancouver, BC, Canada: IEEE, 2023, pp. 10\,597--10\,607.

\bibitem{belghazi2021mine}
\BIBentryALTinterwordspacing
M.~I. Belghazi, A.~Baratin, S.~Rajeswar, S.~Ozair, Y.~Bengio, A.~Courville, and R.~D. Hjelm. (2021-08-14) {{MINE}}: {{Mutual Information Neural Estimation}}. [Online]. Available: \url{http://arxiv.org/abs/1801.04062}
\BIBentrySTDinterwordspacing

\end{thebibliography}
\end{document}